\documentclass[10pt,twocolumn,letterpaper]{article}

\usepackage[pagenumbers]{cvpr} 

\usepackage{booktabs}
\usepackage{url}                  
\usepackage{booktabs}       
\usepackage{amsfonts}       
\usepackage{nicefrac}         
\usepackage{microtype}      
\usepackage[table,xcdraw,dvipsnames]{xcolor} 
\usepackage{caption}
\usepackage{subcaption}
\usepackage{multirow}
\usepackage{caption}
\usepackage{listings}
\usepackage[ruled,vlined]{algorithm2e}
\usepackage{array}
\usepackage{makecell}

\usepackage{amsmath,amssymb} 
\usepackage{color}
\usepackage{multirow}
\usepackage{array}
\usepackage{xspace}
\usepackage{wrapfig}
\usepackage{graphicx,graphics}
\usepackage{mathtools}
\usepackage{pifont}
\usepackage{subcaption}
\usepackage{colortbl}
\usepackage{soul}
\usepackage{enumitem}
\usepackage{multicol}
\usepackage{float}

\usepackage{setspace}
\usepackage{cuted}

\usepackage[pagebackref,breaklinks,colorlinks]{hyperref}

\definecolor{myyellow}{RGB}{255, 250, 205}
\definecolor{myblue}{RGB}{68, 114, 196}
\definecolor{myred}{RGB}{220, 20, 60}
\definecolor{mygreen}{RGB}{46, 139, 87}
\definecolor{myorange}{RGB}{237, 125, 49}
\definecolor{dpurple}{RGB}{128, 0, 128}
\definecolor{lpurple}{RGB}{235, 232, 242}
\definecolor{lblue}{rgb}{0.94, 0.96, 1.0}
\definecolor{grey}{RGB}{128, 128, 128}
\definecolor{lred}{RGB}{255,114,118}

\definecolor{mypurple}{RGB}{126, 116, 212}

\sethlcolor{lblue}

\hypersetup{
    colorlinks=true,
    linkcolor=myred,
    filecolor=myblue,      
    urlcolor=myblue,
    citecolor=mygreen,
}

\usepackage[capitalize]{cleveref}
\crefname{section}{Sec.}{Secs.}
\Crefname{section}{Section}{Sections}
\Crefname{table}{Table}{Tables}
\crefname{table}{Tab.}{Tabs.}

\def\confName{CVPR}
\def\confYear{2023}

\def\methodshort{Cube R-CNN\xspace}
\def\method{Cube R-CNN\xspace}
\def\dataset{\textsc{Omni3D}\xspace}
\def\datasetin{\textsc{Omni3D}$_\textsc{IN}$\xspace}
\def\datasetout{\textsc{Omni3D}$_\textsc{OUT}$\xspace}

\def\APout{AP$^\texttt{OUT}_\threed$\xspace}
\def\APin{AP$^\texttt{IN}_\threed$\xspace}

\def\eg{\emph{e.g.}\xspace}

\def\etal{\emph{et al}\xspace}
\def\etc{\emph{etc.}\xspace}

\def\iou{\texttt{IoU}\xspace}

\def\twod{\text{2D}}
\def\threed{\text{3D}}

\newcommand{\mypar}[1]{\vspace{1mm}\noindent\textbf{#1}}

\newcommand{\myparr}[1]{\vspace{1mm}\noindent\emph{#1}}

\newcolumntype{L}[1]{>{\raggedright\arraybackslash}p{#1}}
\newcolumntype{C}[1]{>{\centering\arraybackslash}p{#1}}
\newcommand{\no}{\textcolor{lred}{\ding{55}}}
\newcommand{\yes}{\textcolor{mygreen}{\ding{51}}}

\newcolumntype{Z}[1]{>{\columncolor{lblue}\centering\arraybackslash}p{#1}}
\newcolumntype{S}[1]{>{\columncolor{lpurple}\centering\arraybackslash}p{#1}}

\newcommand{\vd}[1]{\small{\color{dpurple}{}{$_{#1}$}}}
\newcommand{\vdp}{\phantom{\small{\color{grey}{}{$_{+0.0}$}}}}

\begin{document}

\title{\dataset: A Large Benchmark and Model for 3D Object Detection in the Wild}

\author{%
  Garrick Brazil$^{1}$ \quad 
  Abhinav Kumar$^{2}$ \quad 
  Julian Straub$^{1}$ \quad
  Nikhila Ravi$^{1}$ \quad \smallskip \\
  Justin Johnson$^{1}$ \quad 
  Georgia Gkioxari$^{3}$ \smallskip \smallskip\\
  $^{1}$Meta AI \quad 
  $^{2}$Michigan State University \quad
  $^{3}$Caltech \smallskip\\
}

\makeatletter
\g@addto@macro\@maketitle{
  \vspace{-6mm}
  \begin{figure}[H]
  \setlength{\linewidth}{\textwidth}
  \setlength{\hsize}{\textwidth}
  \centering
  \includegraphics[width=\linewidth]{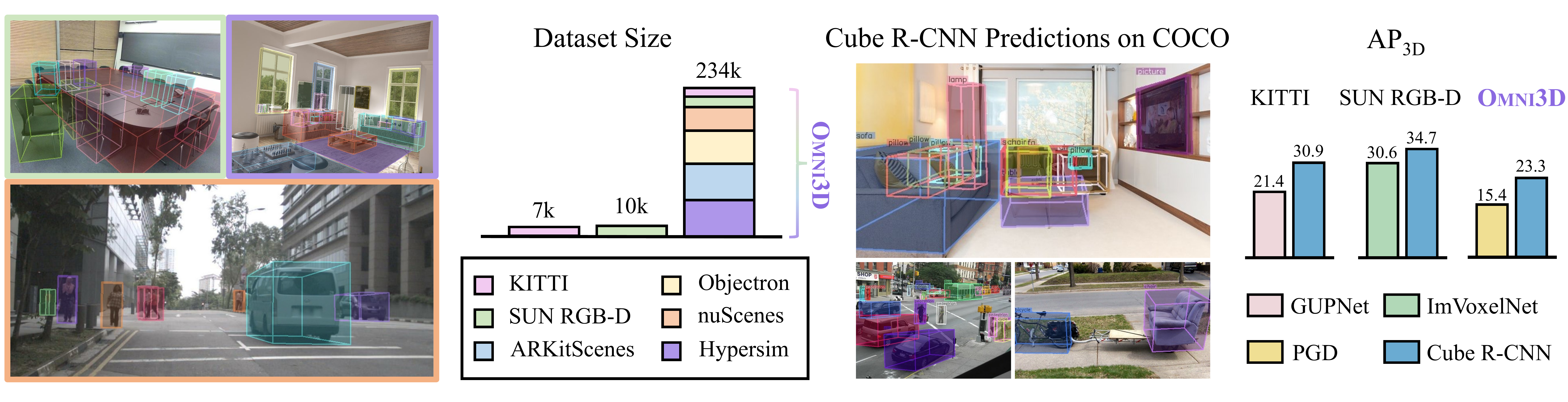}
  \vspace{-8mm}
  \caption{\textbf{Left}: We introduce \dataset, a benchmark for 3D object detection which is larger and more diverse than popular 3D benchmarks. \textbf{Right}: We propose \method, which generalizes to unseen datasets (\eg COCO~\cite{lin2014microsoft}) and outperforms prior works on existing datasets.}\vspace{3mm}
  \label{fig:teaser}
  \end{figure}
}

\maketitle

\begin{abstract}
Recognizing scenes and objects in 3D from a single image is a longstanding goal of computer vision with applications in robotics and AR/VR.
For 2D recognition, large datasets and scalable solutions have led to unprecedented advances.
In 3D, existing benchmarks are small in size and approaches specialize in few object categories and specific domains, \eg urban driving scenes.
Motivated by the success of 2D recognition, we revisit the task of 3D object detection by introducing a large benchmark, called \dataset.
\dataset re-purposes and combines existing datasets resulting in 234k images annotated with more than 3 million instances and 98 categories.
3D detection at such scale is challenging due to variations in camera intrinsics and the rich diversity of scene and object types.
We propose a model, called \method, designed to generalize across camera and scene types with a unified approach. 
We show that \method outperforms prior works on the larger \dataset and existing benchmarks.
Finally, we prove that \dataset is a powerful dataset for 3D object recognition and show that it improves single-dataset performance and can accelerate learning on new smaller datasets via pre-training.\footnote{{We release the \dataset benchmark and \method models at\\ \url{https://github.com/facebookresearch/omni3d}}.}
\end{abstract}
\section{Introduction}
\label{sec:intro}

Understanding objects and their properties from single images is a longstanding problem in computer vision with applications in robotics and AR/VR. 
In the last decade, 2D object recognition~\cite{ren2015faster,he2017maskrcnn, tian2019fcos, law2018cornernet, redmon2016you} has made tremendous advances toward predicting objects on the image plane with the help of large datasets~\cite{lin2014microsoft,gupta2019lvis}. 
However, the world and its objects are three dimensional laid out in 3D space.
Perceiving objects in 3D from 2D visual inputs poses new challenges framed by the task of 3D object detection. 
Here, the goal is to estimate a 3D location and 3D extent of each object in an image in the form of a tight oriented 3D bounding box.

Today 3D object detection is studied under two different lenses: for urban domains in the context of autonomous vehicles~\cite{chen2016monocular,mousavian20173d,brazil2019m3d,liu2020smoke,lu2021geometry} or indoor scenes~\cite{factored3dTulsiani17,huang2018cooperative,kulkarni2019relnet,Nie_2020_CVPR}.
Despite the problem formulation being shared, methods 
share little insights between domains. 
Often approaches are tailored to work only for the domain in question.
For instance, urban methods make assumptions about objects resting on a ground plane and model \textit{only} yaw angles for 3D rotation. 
Indoor techniques may use a confined depth range (\eg up to 6m in~\cite{Nie_2020_CVPR}). 
These assumptions are generally not true in the real world. 
Moreover, the most popular benchmarks for image-based 3D object detection are small. 
Indoor SUN RGB-D~\cite{song2015sun} has 10k images, urban KITTI~\cite{Geiger2012CVPR} has 7k images; 2D benchmarks like COCO~\cite{lin2014microsoft} are 20$\times$ larger.

We address the absence of a general large-scale dataset for 3D object detection by introducing a large and diverse 3D benchmark called \dataset.
\dataset is curated from publicly released datasets, SUN RBG-D~\cite{song2015sun}, ARKitScenes~\cite{dehghan2021arkitscenes}, Hypersim~\cite{hypersim}, Objectron~\cite{objectron2021}, KITTI~\cite{Geiger2012CVPR} and nuScenes~\cite{caesar2020nuscenes}, and comprises 234k images with 3 million objects annotated with 3D boxes across 98 categories including \emph{chair, sofa, laptop, table, cup, shoes, pillow, books, car, person,} \etc
Sec.~\ref{sec:dataset} describes the curation process which involves re-purposing the raw data and annotations from the aforementioned datasets, which originally target different applications. 
As shown in Fig.~\ref{fig:teaser}, \dataset is 20$\times$ larger than existing popular benchmarks used for 3D detection, SUN RGB-D and KITTI.
For efficient evaluation on the large \dataset, we introduce a new algorithm for intersection-over-union of 3D boxes which is 450$\times$ faster than previous solutions~\cite{objectron2021}.
We empirically prove the impact of \dataset as a large-scale dataset and show that it improves single-dataset performance by up to 5.3\% AP on urban and 3.8\% on indoor benchmarks.

On the large and diverse \dataset, we design a general and simple 3D object detector, called \method, inspired by advances in 2D and 3D recognition of recent years~\cite{ren2015faster, simonelli2019disentangling, lu2021geometry, gkioxari2019mesh}.
\method detects all objects and their 3D location, size and rotation end-to-end from a single image of any domain and for many object categories.
Attributed to \dataset's diversity, our model shows strong generalization and outperforms prior works for indoor and urban domains with one unified model, as shown in Fig.~\ref{fig:teaser}.
Learning from such diverse data comes with challenges as \dataset contains images of highly varying focal lengths which exaggerate scale-depth ambiguity (Fig.~\ref{fig:virtual_depth}).
We remedy this by operating on \textit{virtual depth} which transforms object depth with the same virtual camera intrinsics across the dataset.
An added benefit of virtual depth is that it allows the use of data augmentations (\eg image rescaling) during training, which is a critical feature for 2D  detection~\cite{wu2019detectron2, mmdetection}, and as we show, also for 3D.
Our approach with one unified design outperforms prior best approaches in AP$_\text{3D}$, ImVoxelNet~\cite{rukhovich2022imvoxelnet} by 4.1\% on indoor SUN RGB-D, GUPNet~\cite{lu2021geometry} by 9.5\% on urban KITTI, and PGD~\cite{wang2022probabilistic} by 7.9\% on \dataset.

We summarize our contributions:
\begin{itemize}
\item We introduce \dataset, a benchmark for image-based 3D object detection sourced from existing 3D datasets, which is 20$\times$ larger than existing 3D benchmarks.
\item We implement a new algorithm for IoU of 3D boxes, which is 450$\times$ faster than prior solutions.
\item We design a general-purpose baseline method, \method, which tackles 3D object detection for many categories and across domains with a unified approach. We propose \emph{virtual depth} to eliminate the ambiguity from varying camera focal lengths in \dataset.
\end{itemize}
\section{Related Work}
\label{sec:related}

\method and \dataset draw from key research advances in 2D and 3D object detection.

\mypar{2D Object Detection.} 
Here, methods include two-stage approaches~\cite{ren2015faster,he2017maskrcnn} which predict object regions with a region proposal network (RPN) and then refine them via an MLP.
Single-stage detectors~\cite{redmon2016you,liu2016ssd,lin2017focal,zhou2019objects, tian2019fcos} omit the RPN and predict regions directly from the backbone. 

\mypar{3D Object Detection.}
Monocular 3D object detectors predict 3D cuboids from single input images. 
There is extensive work in the urban self-driving domain where the \emph{car} class is at the epicenter \cite{liu2021deep, reading2021categorical, chen2016monocular, mousavian20173d, liu2021ground, wang2021progressive, FuCVPR18DORN, gu2022homography, zhou2020iafa, ma2019accurate, he2021aug3d, simonelli2020towards, simonelli2019disentangling, huang2022monodtr,wang2021fcos3d}.
CenterNet~\cite{zhou2019objects} predicts 3D depth and size from fully-convolutional center features, and is extended by~\cite{liu2020smoke, lu2021geometry, liu2021learning, zhang2021objects, zhou2021monocular, chen2020monopair, liu2021autoshape, li2020rtm3d, Ma_2021_CVPR, zhou2020iafa,kumar2022deviant}. 
M3D-RPN~\cite{brazil2019m3d} trains an RPN with 3D anchors, enhanced further by~\cite{ding2020learning, brazil2020kinematic, zou2021devil, wang2021depth, kumar2021groomed}.
FCOS3D~\cite{wang2021fcos3d} extends the anchorless FCOS~\cite{tian2019fcos} detector to predict 3D cuboids. 
Its successor PGD~\cite{wang2022probabilistic} furthers the approach with probabilistic depth uncertainty.
Others use pseudo depth~\cite{bao2019monofenet, you2019pseudo, wang2019pseudo, chen2022pseudo, park2021dd3d, Ma_2020_ECCV} and explore depth and point-based LiDAR techniques~\cite{qi2018frustum, ku2018joint}.
Similar to ours,~\cite{chen2021monorun, simonelli2019disentangling, simonelli2020towards} add a 3D head, specialized for urban scenes and objects, on two-stage Faster R-CNN.
\cite{he2021aug3d, simonelli2020towards} augment their training by synthetically generating depth and box-fitted views, coined as virtual views or depth.
In our work, virtual depth aims at addressing \textit{varying focal lengths}.  

For indoor scenes, a vast line of work tackles room layout estimation~\cite{hedau2009recovering,lee2009geometric,mallya2015learning,dasgupta2016delay}.
Huang \etal~\cite{huang2018cooperative} predict 3D oriented bounding boxes for indoor objects.
Factored3D~\cite{factored3dTulsiani17} and 3D-RelNet~\cite{kulkarni2019relnet} jointly predict object voxel shapes. 
Total3D~\cite{Nie_2020_CVPR} predicts 3D boxes and meshes by additionally training on datasets with annotated 3D shapes.
ImVoxelNet~\cite{rukhovich2022imvoxelnet} proposes domain-specific methods which share an underlying framework for processing volumes of 3D voxels. 
In contrast, we explore 3D object detection in its general form by tackling outdoor and indoor domains jointly in a single model and with a vocabulary of $5\times$ more categories. 

\mypar{3D Datasets.}
KITTI~\cite{Geiger2012CVPR} and SUN RGB-D~\cite{song2015sun} are popular datasets for 3D object detection on urban and indoor scenes respectively.
Since 2019, 3D datasets have emerged, both for indoor~\cite{objectron2021,hypersim,dehghan2021arkitscenes, dai2017scannet, avetisyan2019scan2cad} and outdoor~\cite{sun2020scalability,caesar2020nuscenes,houston2020one, chang2019argoverse, lyft,huang2019apolloscape,gahlert2020cityscapes}.
In isolation, these datasets target different tasks and applications and have unique properties and biases, \eg object and scene types, focal length, coordinate systems, \etc
In this work, we unify existing representative datasets~\cite{Geiger2012CVPR, song2015sun, objectron2021,hypersim,dehghan2021arkitscenes, caesar2020nuscenes}.
We process the raw visual data, re-purpose their annotations, and carefully curate the union of their semantic labels in order to build a coherent large-scale benchmark, called \dataset.
\dataset is $20\times$ larger than widely-used benchmarks and notably more diverse.
As such, new challenges arise stemming from the increased variance in visual domain, object rotation, size, layouts, and camera intrinsics.

\begin{figure*}[t]
	{\small \hspace{3mm} \dataset  \hspace{0.2mm}  SUN RGB-D  \hspace{0.5mm}  KITTI \hspace{6mm} COCO \hspace{5.5mm} LVIS} \\
	\includegraphics[width=0.98\textwidth]{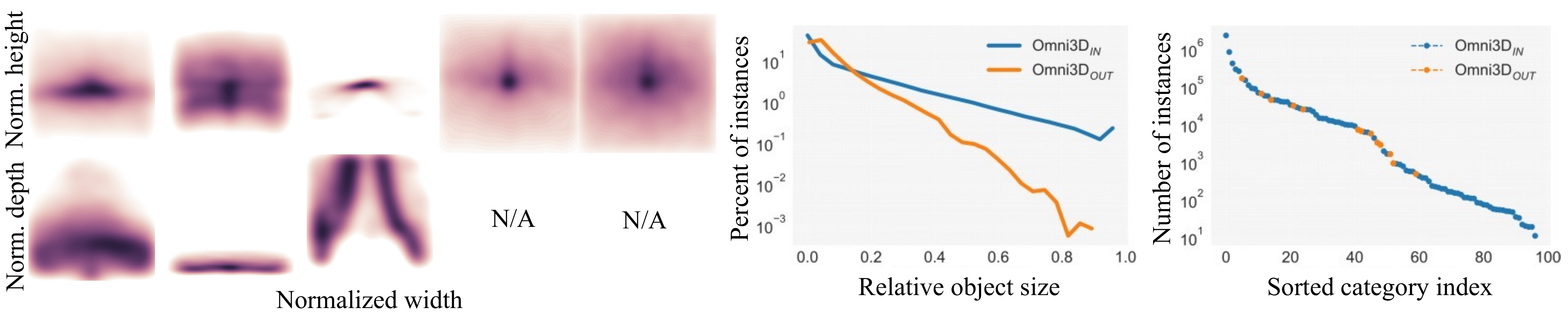}
	\begin{minipage}{0.40\textwidth} \centering \hspace{10mm} \vspace{-0.5mm} (a) Spatial statistics \end{minipage}
   	\begin{minipage}{0.33\textwidth} \centering \hspace{6mm} (b) 2D Scale statistics \end{minipage}
	\begin{minipage}{0.24\textwidth} \centering \hspace{0mm} (c) Category statistics \end{minipage}
	\vspace{-2mm}
     	\caption{\dataset analysis. (a) distribution of object centers on (top) normalized image, XY-plane, and (bottom) normalized depth, XZ-plane, (b) relative 2D object scales, and (c) category frequency. 
    }
	\label{fig:dataset_stats}
	\vspace{-3mm}
\end{figure*}

\section{The \dataset Benchmark}
\label{sec:dataset}

The primary benchmarks for 3D object detection are small, focus on a few categories and are of a single domain.
For instance, the popular KITTI \cite{Geiger2012CVPR} contains only urban scenes, has 7k images and 3 categories, with a focus on \texttt{car}.
SUN RGB-D~\cite{song2015sun} has 10k images.
The small size and lack of variance in 3D datasets is a stark difference to 2D counterparts, such as COCO~\cite{lin2014microsoft} and LVIS~\cite{gupta2019lvis}, which have pioneered progress in 2D recognition. 

We aim to bridge the gap to 2D by introducing \dataset, a large-scale and diverse benchmark for image-based 3D object detection consisting of 234k images, 3 million labeled 3D bounding boxes, and 98 object categories. 
We source from recently released 3D datasets of urban (nuScenes~\cite{caesar2020nuscenes} and KITTI~\cite{Geiger2012CVPR}), indoor (SUN RGB-D~\cite{song2015sun}, ARKitScenes~\cite{dehghan2021arkitscenes} and Hypersim~\cite{hypersim}), and general (Objectron~\cite{objectron2021}) scenes.
Each of these datasets target different applications (\eg point-cloud recognition or reconstruction in~\cite{dehghan2021arkitscenes}, inverse rendering in~\cite{hypersim}), provide visual data in different forms (\eg videos in~\cite{objectron2021,dehghan2021arkitscenes}, rig captures in~\cite{caesar2020nuscenes}) and annotate different object types.
To build a coherent benchmark, we process the varying raw visual data, re-purpose their annotations to extract 3D cuboids in a \emph{unified 3D camera coordinate system}, and carefully curate the final vocabulary.
More details about the benchmark creation in the Appendix.

We analyze \dataset and show its rich spatial and semantic properties proving it is visually diverse, similar to 2D data, and highly challenging for 3D as depicted in Fig.~\ref{fig:dataset_stats}.
We show the value of \dataset for the task of 3D object detection with extensive quantitative analysis in Sec.~\ref{sec:exp}.

\subsection{Dataset Analysis}

\mypar{Splits.}
We split the dataset into 175k/19k/39k images for train/val/test respectively, consistent with original splits when available, and otherwise free of overlapping video sequences in splits.  
We denote indoor and outdoor subsets as \datasetin (SUN RGB-D, Hypersim, ARKit), and \datasetout (KITTI, nuScenes).
Objectron, with primarily close-up objects, is used only in the \textit{full} \dataset setting.

\mypar{Layout statistics.}
Fig.~\ref{fig:dataset_stats}(a) shows the distribution of object centers onto the image plane by projecting centroids on the XY-plane (top row), and the distribution of object depths by projecting centroids onto the XZ-plane (bottom row).
We find that \dataset's spatial distribution has a center bias, similar to 2D datasets COCO and LVIS.
Fig.~\ref{fig:dataset_stats}(b) depicts the relative object size distribution, defined as the square root of object area divided by image area.
Objects are more likely to be small in size similar to LVIS (Fig. 6c in~\cite{gupta2019lvis}) suggesting that \dataset is also challenging for 2D detection, while objects in \datasetout are noticeably smaller.

The bottom row of Fig.~\ref{fig:dataset_stats}(a) normalizes object depth in a $[0, 20\text{m}]$ range, chosen for visualization and satisfies $88\%$ of object instances; \dataset depth ranges as far as $300$m.
We observe that the \dataset depth distribution is far more diverse than SUN RGB-D and KITTI, which are biased toward near or road-side objects respectively.
See Appendix for each data source distribution plot. 
Fig.~\ref{fig:dataset_stats}(a) demonstrates \dataset's rich diversity in spatial distribution and depth which suffers significantly less bias than existing 3D benchmarks and is comparable in complexity to 2D datasets.

\mypar{2D and 3D correlation.}
A common assumption in urban scenes is that objects rest on a ground plane and appear smaller with depth. 
To verify if that is true generally, we compute correlations. 
We find that 2D $y$ and 3D $z$ are indeed fairly correlated in \datasetout at $0.524$, but \textit{significantly} less in \datasetin at $0.006$.
Similarly, relative 2D object size ($\sqrt{h \cdot w}$) and $z$ correlation is $0.543$ and $0.102$ respectively.
This confirms our claim that common assumptions in urban scenes are not generally true, making the task challenging.

\mypar{Category statistics.}
 Fig.~\ref{fig:dataset_stats}(c) plots the distribution of instances across the 98 categories of \dataset.
The long-tail suggests that low-shot recognition in both 2D and 3D will be critical for performance. 
In this work, we want to focus on large-scale and diverse 3D recognition, which is comparably unexplored.
We therefore filter and focus on categories with at least $1000$ instances.
This leaves 50 diverse categories including \emph{chair, sofa, laptop, table, books, car, truck, pedestrian} and more, 19 of which have more than 10k instances.
We provide more per-category details in the Appendix.

\begin{figure*}[t!]
\centering
\includegraphics[width=1.0\linewidth]{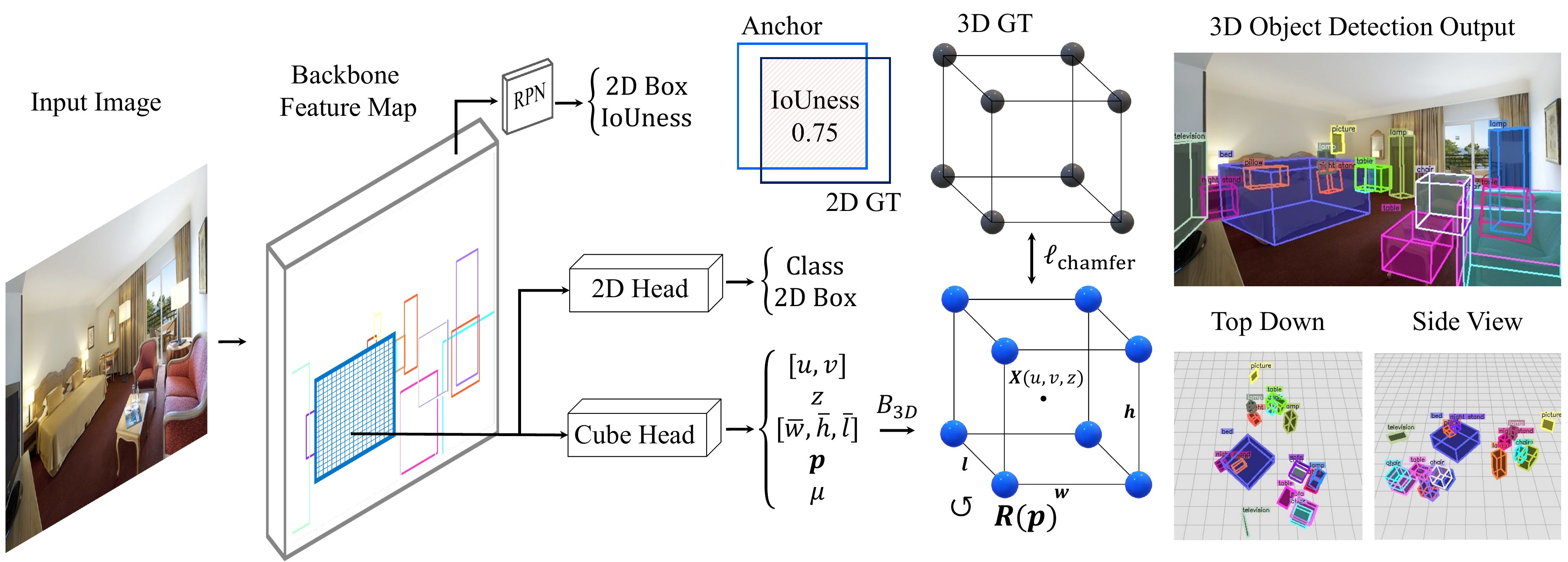}
\vspace{-7mm}
\caption{Overview. \method takes as input an RGB image, detects all objects in 2D and predicts their 3D cuboids $B_\text{3D}$. During training, the 3D corners of the cuboids are compared against 3D ground truth with the point-cloud chamfer distance.}
\label{fig:overview}
\vspace{-2mm}
\end{figure*}

\section{Method}
\label{sec:method}

Our goal is to design a simple and effective model for \textit{general} 3D object detection.
Hence, our approach is free of domain or object specific strategies.
We design our 3D object detection framework by extending Faster R-CNN~\cite{ren2015faster} with a 3D object head which predicts a cuboid per each detected 2D object.
We refer to our method as \emph{\method}.
Figure~\ref{fig:overview} shows an overview of our approach.

\subsection{\method}

Our model builds on Faster R-CNN~\cite{ren2015faster}, an end-to-end region-based object detection framework.
Faster-RCNN consists of a backbone network, commonly a CNN, which embeds the input image into a higher-dimensional feature space.
A region proposal network (RPN) predicts regions of interest (RoIs) representing object candidates in the image.
A 2D box head inputs the backbone feature map and processes each RoI to predict a category and a more accurate 2D box.
Faster R-CNN can be easily extended to tackle more tasks by adding task-specific heads, \eg Mask R-CNN~\cite{he2017maskrcnn} adds a mask head to additionally output object silhouettes.

For the task of 3D object detection, we extend Faster R-CNN by introducing a 3D detection head which predicts a 3D cuboid for each detected 2D object.
\method extends Faster R-CNN in three ways:
(a) we replace the binary classifier in RPN which predicts region \emph{objectness} with a regressor that predicts \emph{IoUness},
(b) we introduce a \emph{cube head} which estimates the parameters to define a 3D cuboid for each detected object (similar in concept to~\cite{simonelli2019disentangling}), and (c) we define a new training objective which incorporates a \emph{virtual depth} for the task of 3D object detection.

\mypar{IoUness.}
The role of RPN is two-fold: (a) it proposes RoIs by regressing 2D box coordinates from pre-computed anchors and (b) it classifies regions as \emph{object or not} (objectness). 
This is sensible in exhaustively labeled datasets where it can be reliably assessed if a region contains an object.
However, \dataset combines many data sources with no guarantee that all instances of all classes are labeled.
We overcome this by replacing objectness with \emph{IoUness}, applied only to foreground.
Similar to~\cite{kim2022learning}, a regressor predicts IoU between a RoI and a ground truth.
Let $o$ be the predicted IoU for a RoI and $\hat{o}$ be the 2D IoU between the region and its ground truth; we apply a binary cross-entropy (CE) loss $\mathcal{L}_\text{IoUness} = \ell_\text{CE}(o, \hat{o})$.
We train on regions whose IoU exceeds 0.05 with a ground truth in order to learn IoUness from a wide range of region overlaps. 
Thus, the RPN training objective becomes $\mathcal{L}_\text{RPN} = \hat{o} \cdot (\mathcal{L}_\text{IoUness} + \mathcal{L}_\text{reg})$, where $\mathcal{L}_\text{reg}$ is the 2D box regression loss from~\cite{ren2015faster}. 
The loss is weighted by $\hat{o}$ to prioritize candidates close to true objects.

\mypar{Cube Head.}
We extend Faster R-CNN with a new head, called cube head, to predict a 3D cuboid for each detected 2D object.
The cube head inputs $7\times7$ feature maps pooled from the backbone for each predicted region and feeds them to $2$ fully-connected (FC) layers with $1024$ hidden dimensions.
All 3D estimations in the cube head are category-specific.
The cube head represents a 3D cuboid with 13 parameters each predicted by a final FC layer:
\begin{itemize}[noitemsep, nolistsep]
\item $[u, v] $ represent the projected 3D center on the image plane relative to the 2D RoI
\item $z \in \mathbb{R}_+$ is the object's center depth in meters transformed from virtual depth $z_v$ (explained below)
\item $[\bar{w}, \bar{h}, \bar{l}] \in \mathbb{R}_+^3$ are the log-normalized physical box dimensions in meters
\item  $\mathbf{p} \in \mathbb{R}^6$ is the continuous 6D~\cite{zhou2019continuity} allocentric rotation 
\item  $\mu \in \mathbb{R_+}$ is the predicted 3D uncertainty
\end{itemize}

The above parameters form the final 3D box in camera view coordinates for each detected 2D object.
The object's 3D center $\mathbf{X}$ is estimated from the predicted 2D projected center $[u, v]$ and depth $z$ via 
\begin{equation}
  \small
  \mathbf{X}(u, v, z) = \begin{pmatrix} \frac{z}{f_x}(r_x + u r_w - p_x)\,, \frac{z}{f_y} (r_y + v  r_h - p_y) ,\, z \end{pmatrix}
\end{equation} 
where $[r_x, r_y, r_w, r_h]$ is the object's 2D box, $(f_x, f_y)$ are the camera's known focal lengths and $(p_x, p_y)$ the principal point.
The 3D box dimensions $\mathbf{d}$ are derived from $[\bar{w}, \bar{h}, \bar{l}]$ which are log-normalized with category-specific pre-computed means $(w_0, h_0, l_0)$ for width, height and length respectively, and are are arranged into a diagonal matrix via 
\begin{equation}
  \small
  \mathbf{d}(\bar{w}, \bar{h}, \bar{l}) = \text{diag}\begin{pmatrix} \exp(\bar{w}) 
  w_0,\, \exp(\bar{h}) h_0,\, \exp(\bar{l}) l_0 \end{pmatrix}
\end{equation}
Finally, we derive the object's pose $\mathbf{R}(\mathbf{p})$ as a $3 \times 3$ rotation matrix based on a 6D parameterization (2 directional vectors) of $\mathbf{p}$ following~\cite{zhou2019continuity} which is converted from allocentric to egocentric rotation similar to~\cite{kundu20183d}, defined formally in Appendix. 
The final 3D cuboid, defined by $8$ corners, is
\begin{equation}
\small
B_\textrm{3D}(u, v, z, \bar{w}, \bar{h}, \bar{l}, \mathbf{p}) = \mathbf{R}(\mathbf{p}) \, \mathbf{d}(\bar{w}, \bar{h}, \bar{l}) \, B_\text{unit} + \mathbf{X}(u, v, z)
\label{eq:box3d}
\end{equation}
where $B_\text{unit}$ are the 8 corners of an axis-aligned unit cube centered at $(0,0,0)$.
Lastly, $\mu$ denotes a learned 3D uncertainty, which is mapped to a confidence at inference then joined with the classification score $s$ from the 2D box head to form the final score for the prediction, as $\sqrt{s \cdot \exp(-{\mu})}$.

\mypar{Training objective.}
Our training objective consists of 2D losses from the RPN and 2D box head and 3D losses from the cube head.
The 3D objective compares each predicted 3D cuboid with its matched ground truth via a chamfer loss, treating the 8 box corners of the 3D boxes as point clouds, namely $\mathcal{L}_\text{3D}^\text{all} = \ell_\text{chamfer}(B_\text{3D}, B^\text{gt}_\text{3D})$.
Note that $\mathcal{L}_\text{3D}^\text{all}$ \textit{entangles} all 3D variables via the box predictor $B_\text{3D}$ (Eq.~\ref{eq:box3d}), such that that errors in variables may be ambiguous from one another. 
Thus, we isolate each variable group with separate disentangled losses, following~\cite{simonelli2019disentangling}.
The disentangled loss for each variable group substitutes all but its variables with the ground truth from Eq.~\ref{eq:box3d} to create a pseudo box prediction.\linebreak
For example, the disentangled loss for the projected center $[u, v]$ produces a 3D box with all but $(u, v)$ replaced with the \textit{true} values and then compares to the ground truth box,
\begin{equation}
\mathcal{L}_\text{3D}^{(u,v)} = \| B_\text{3D}(u, v, z^\text{gt}, \bar{w}^\text{gt}, \bar{h}^\text{gt}, \bar{l}^\text{gt}, \mathbf{p}^\text{gt}) - B_\text{3D}^\text{gt}\|_1
\end{equation}
We use an $L_1$ loss for $\mathcal{L}_\text{3D}^{(u,v)}$, $\mathcal{L}_\text{3D}^{(z)}$ and $\mathcal{L}_\text{3D}^{(\bar{w}, \bar{h}, \bar{l})}$ and chamfer for $\mathcal{L}_\text{3D}^\mathbf{p}$ to account for cuboid symmetry such that rotation matrices of Euler angles modulo $\pi$ produce the same non-canonical 3D cuboid.
Losses in box coordinates have natural advantages over losses which directly compare variables to ground truths in that their gradients are appropriately weighted by the error as shown in~\cite{simonelli2019disentangling}. The 3D objective is defined as, $\mathcal{L}_\text{3D} = \mathcal{L}_\text{3D}^{(u,v)} + \mathcal{L}_\text{3D}^{(z)} + \mathcal{L}_\text{3D}^{(\bar{w}, \bar{h}, \bar{l})} + \mathcal{L}_\text{3D}^\mathbf{p} + \mathcal{L}_\text{3D}^\text{all}$.

The collective training objective of \method is,
\begin{equation}
\mathcal{L} = \mathcal{L}_\text{RPN} + \mathcal{L}_\text{2D} + \sqrt{2} \cdot \exp(-\mu) \cdot \mathcal{L}_\text{3D}  + \mu
\label{eq:loss}
\end{equation}
$\mathcal{L}_\text{RPN}$ is the RPN loss, described above, and $\mathcal{L}_\text{2D}$ is the 2D box head loss from~\cite{ren2015faster}.
The 3D loss is weighted by the predicted 3D uncertainty (inspired by~\cite{lu2021geometry}), such that the model may trade a penalty to reduce the 3D loss when uncertain. 
In practice, $\mu$ is helpful for both improving the 3D ratings at inference and reducing the loss of hard samples in training. 

\begin{figure}[t]
   \centering
   \includegraphics[width=\linewidth]{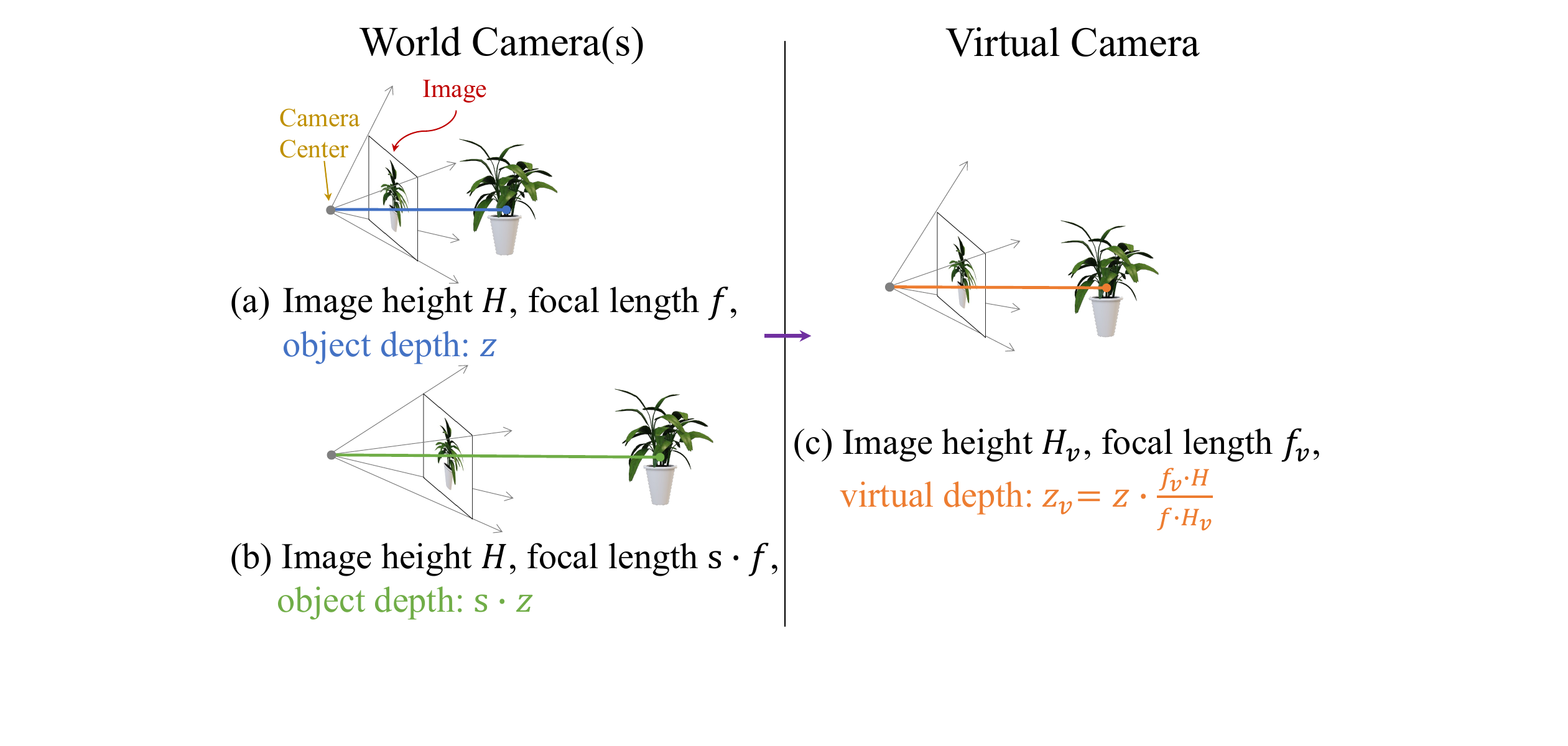}
    \vspace{-7mm}
   \caption{Varying camera intrinsics exaggerate scale-depth ambiguity -- the same object at two different depths can project to the same image, as in (a) and (b). 
  We address this by introducing a \emph{virtual camera} which is invariant to intrinsics and transforms the object's depth to a \emph{virtual depth} $z_v$ such that the effective image size $H_v$ and focal length $f_v$ are consistent across images.}
   \label{fig:virtual_depth}
    \vspace{-4mm}
\end{figure}

\subsection{Virtual Depth}

A critical part of 3D object detection is predicting an object's depth in metric units. 
Estimating depth from visual cues requires an implicit mapping of 2D pixels to 3D distances, which is more ambiguous if camera intrinsics vary.
Prior works are able to ignore this as they primarily train on images from a single sensor. 
\dataset contains images from many sensors and thus demonstrates large variations in camera intrinsics.
We design \method to be robust to intrinsics by predicting the object's \textit{virtual depth} $z_v$, which projects the metric depth $z$ to an invariant camera space.

Virtual depth scales the depth using the (known) camera intrinsics such that the effective image size and focal length are \textit{consistent across the dataset}, illustrated in Fig.~\ref{fig:virtual_depth}.
The \textit{effective} image properties are referred to as virtual image height $H_v$ and virtual focal length $f_v$, and are both hyperparameters.
If $z$ is the true metric depth of an object from an image with height $H$ and focal length $f$, its virtual depth is defined as $z_v = z \cdot \frac{f_v}{f}  \frac{H}{H_v}$.
The derivation is in the Appendix.

Invariance to camera intrinsics via virtual depth \textit{also} enables scale augmentations during training, since $H$ can vary without harming the consistency of $z_v$. 
Data augmentations from image resizing tend to be critical for 2D models but are not often used in 3D methods~\cite{brazil2019m3d, liu2020smoke, simonelli2019disentangling} since if unaccounted for they increase scale-depth ambiguity.
Virtual depth lessens such ambiguities and therefore enables powerful data augmentations in training. 
We empirically prove the two-fold effects of \textit{virtual depth} in Sec.~\ref{sec:exp}

\begin{figure*}[t]
\centering
\includegraphics[width=0.99\linewidth]{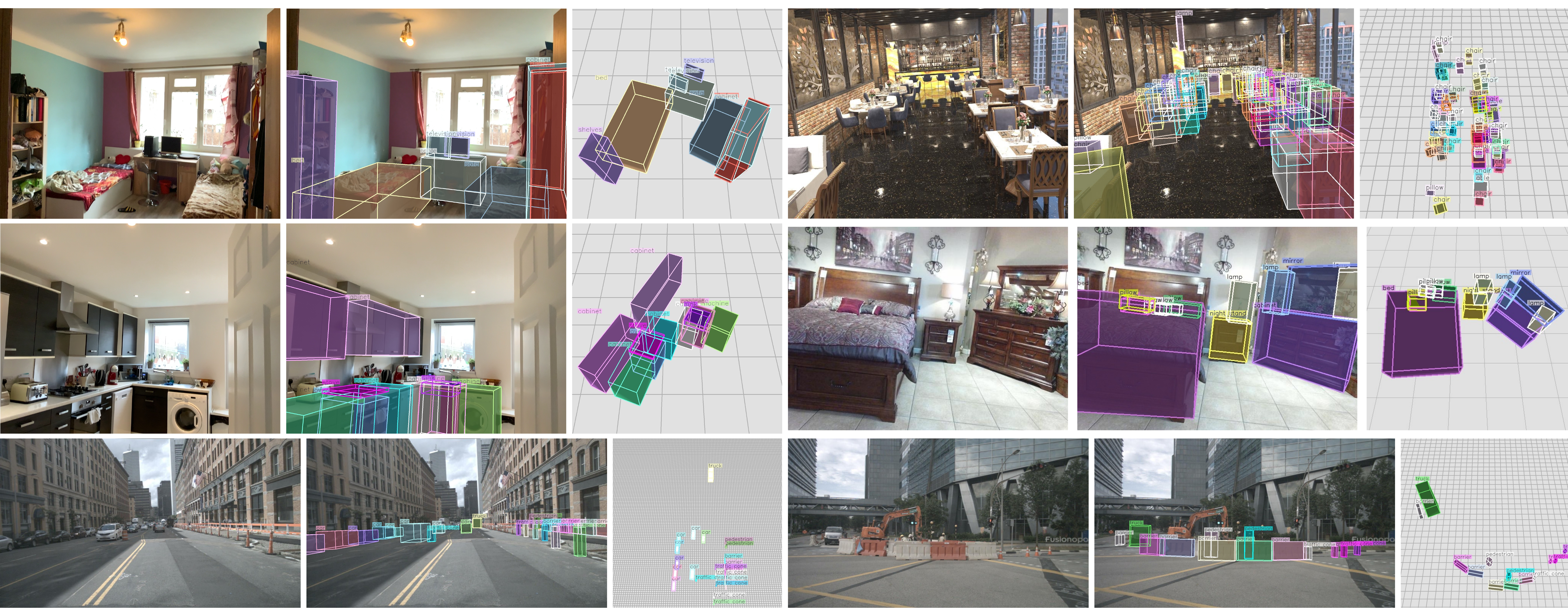}
\vspace{-3mm}
\caption{\method predictions on \dataset test. For each example, we show the input image, the 3D predictions overlaid on the image and a top view with a base composed of 1m$\times$1m tiles. See more results from images and videos in the Appendix.}
\label{fig:qualitative}
\vspace{-3mm}
\end{figure*}

\section{Experiments}
\label{sec:exp}

We tackle image-based 3D object detection on \dataset and compare to prior best methods.
We evaluate \method on existing 3D benchmarks with a single unified model.
Finally, we prove the effectiveness of \dataset as a large-scale 3D object detection dataset by showing comprehensive cross-dataset generalization and its impact for pre-training.

\mypar{Implementation details.}
We implement \method using Detectron2~\cite{wu2019detectron2} and PyTorch3D~\cite{ravi2020pytorch3d}. 
Our backbone is DLA34~\cite{yu2018deep}-FPN~\cite{lin2017feature} pretrained on ImageNet~\cite{russakovsky2015imagenet}.
We train for 128 epochs with a batch size of 192 images across 48 V100s. 
We use SGD with a learning rate of 0.12 which decays by a factor of 10 after 60\% and 80\% of training.
During training, we use random data augmentation of horizontal flipping and scales $\in [0.50,~1.25]$, enabled by virtual depth.
Virtual camera parameters are set to $f_v = H_v = 512$.
Source code and models are publicly available.

\mypar{Metric.}
Following popular benchmarks for 2D and 3D recognition, we use average-precision (AP) as our 3D metric.
Predictions are matched to ground truth by measuring their overlap using $\text{\iou}_\threed$ which computes the intersection-over-union of 3D cuboids.
We compute a \emph{mean} AP$_\threed$ across all 50 categories in \dataset and over a range of $\text{\iou}_\threed$ thresholds $ \tau \in [0.05, 0.10,\dots, 0.50]$. 
The range of $\tau$ is more relaxed than in 2D to account for the new dimension -- see the Appendix for a comparison between 3D and 2D ranges for IoU.
General 3D objects may be occluded by other objects or truncated by the image border, and can have arbitrarily small projections.
Following~\cite{Geiger2012CVPR}, we ignore objects with high occlusion ($>66\%$) or truncation ($>33\%$), and tiny projected objects ($<6.25\%$ image height).
Moreover, we also report AP$_\threed$ at varying levels of depth $d$ as near: $0< d \leq 10m$, medium: $10m < d \leq 35m$, far: $35m < d \leq \small{\infty}$.

\mypar{Fast $\text{\iou}_\threed$.}
$\text{\iou}_\threed$ compares two 3D cuboids by computing their intersection-over-union.
Images usually have many objects and produce several predictions so $\text{\iou}_\threed$ computations need to be fast. 
Prior implementations~\cite{Geiger2012CVPR} approximate $\text{\iou}_\threed$ by projecting 3D boxes on a ground plane and multiply the top-view 2D intersection with the box heights to compute a 3D volume.
Such approximations become notably inaccurate when objects are not on a planar ground or have an arbitrary orientation (\eg with nonzero pitch or roll).\linebreak
Objectron~\cite{objectron2021} provides an exact solution, but relies on external libraries~\cite{virtanen2020scipy,barber1996quickhull} implemented in C++ and is not batched.
We implement a new, fast and exact algorithm which computes the intersecting shape by representing cuboids as meshes and finds face intersections -- described in Appendix.
Our algorithm is batched with C++ and CUDA support.
Our implementation is $90\times$ faster than Objectron in C++, and $450\times$ faster in CUDA.
As a result, evaluation on the large \dataset takes only seconds instead of several hours.

\subsection{Model Performance}
\label{sec:ablations}

First, we ablate the design choices of \method.
Then, we compare to state-of-the-art methods on existing benchmarks and \dataset and show that it performs superior or on par with prior works which are specialized for their respective domains.
Our \method uses a single unified design to tackle general 3D object detection across domains.
Fig.~\ref{fig:qualitative} shows qualitative predictions on the \dataset test set.

\begin{table*}[t!]
\centering
\scalebox{0.94}{
\begin{tabular}{L{.17\textwidth}|C{.05\textwidth}C{.05\textwidth}C{.05\textwidth}C{.05\textwidth}C{.05\textwidth}C{.05\textwidth}|C{.05\textwidth}C{.05\textwidth}}
\method & AP$_\threed$  & AP$_\threed^\text{25}$ &AP$_\threed^\text{50}$ & AP$_\threed^\textrm{near}$ & AP$_\threed^\textrm{med}$ & AP$_\threed^\textrm{far}$  & \APin & \APout \\
\hline
\emph{w/o} disentangled 	& 22.6 & 24.3 & 9.2 & 26.8 & 11.8 & 8.2 & 14.5 & 30.9  \\
\emph{w/o} IoUness 			& 22.2 & 23.6 & 8.8 & 26.4 & 11.1 & 8.3 & 14.3 & 31.0  \\
\emph{w/o} $\mathcal{L}_\text{3D}^\text{all}$ 	& 20.2 & 21.8 & 6.4 & 26.4 & \bf{12.1} & 7.4 & 13.8 & 29.1  \\
\emph{w/o} scale aug. 		& 20.2 & 21.5 & 8.0 & 23.5 & ~~9.8 & 6.8 & 12.3 & 28.1  \\
\emph{w/o} scale $|$ virtual 		& 19.8 & 21.2 & 7.5 & 23.4& ~~8.6 & 5.7 & 12.2 & 26.0  \\ 
\emph{w/o} virtual depth 		& 17.3 & 18.4 & 4.4 & 22.7& ~~7.9 & 7.1 & 13.2 & 30.3  \\ 
\emph{w/o} uncertainty $\mu$ 	& 17.2 & 18.3 & 5.4 & 20.5 & 10.8 & 7.0 & $~~$9.1 & 25.4 \\
ours 						& \bf{23.3} & \bf{24.9} & \bf{9.5} & \bf{27.9} & \bf{12.1} & \bf{8.5} & \bf{15.0} & \bf{31.9} \\
\end{tabular}
}
\vspace{-3mm}
\caption{\method ablations. We report AP$_\threed$ on \dataset, then at IoU thresholds 0.25 and 0.50, and for near, medium and far objects. We further report AP$_\threed$ when both training and evaluating on \datasetin and \datasetout subsets.}
\label{tab:ablations}
\vspace{-6mm}
\end{table*}

\begin{table}[t]
\centering
\scalebox{0.85}{
\begin{tabular}{L{.13\textwidth}|C{.01\textwidth}|C{.07\textwidth}C{.07\textwidth}C{.07\textwidth}|C{.07\textwidth}}
 & &  \multicolumn{3}{c|}{\datasetout} & \dataset \\
Method & $f$ & AP$^\texttt{KIT}_\threed$ & AP$^\texttt{NU}_\threed$ &  \APout & AP$_\threed$ \\
\hline
M3D-RPN~\cite{brazil2019m3d}			& \no & 10.4\vdp & 17.9\vdp & 13.7\vdp  & -\vdp\\
SMOKE~\cite{liu2020smoke}		 		& \no & 25.4\vdp & 20.4\vdp & 19.5\vdp & \phantom{0}9.6\vdp \\
FCOS3D~\cite{wang2021fcos3d}      		& \no & 14.6\vdp & 20.9\vdp & 17.6\vdp & \phantom{0}9.8\vdp\\
PGD~\cite{wang2022probabilistic}    			& \no & 21.4\vdp & 26.3\vdp & 22.9\vdp & 11.2\vdp\\
\hline
GUPNet~\cite{lu2021geometry}		 		& \yes & 24.5\vdp & 20.5\vdp & 19.9\vdp & -\vdp\\
M3D-RPN \emph{+vc}    					& \yes & 16.2\vd{+5.8} & 20.4\vd{+2.9} & 17.0\vd{+3.3} & -\vdp \\
ImVoxelNet~\cite{rukhovich2022imvoxelnet} 	& \yes & 23.5\vdp & 23.4\vdp & 21.5\vdp & \phantom{0}9.4\vdp \\
SMOKE \emph{+vc}          				& \yes & 25.9\vd{+0.5} & 20.4\vd{+0.0} & 20.0\vd{+0.5} & 10.4\vd{+0.8}\\
FCOS3D \emph{+vc}					& \yes & 17.8\vd{+3.2} & 25.1\vd{+4.2} & 21.3\vd{+3.7} & 10.6\vd{+0.8}\\
PGD \emph{+vc}						& \yes & 22.8\vd{+1.4} & 31.2\vd{+4.9} &  26.8\vd{+3.9} & 15.4\vd{+4.2} \\
\methodshort 							& \yes & \bf{36.0}\vdp & \bf{32.7}\vdp & \bf{31.9}\vdp & \bf{23.3}\vdp \\
\end{tabular}
}
\vspace{-4mm}
\caption{\dataset comparison with GUPNet~\cite{lu2021geometry}, ImVoxelNet~\cite{rukhovich2022imvoxelnet}, M3D-RPN~\cite{brazil2019m3d}, FCOS3D~\cite{wang2021fcos3d}, PGD~\cite{wang2022probabilistic}, SMOKE~\cite{liu2020smoke}. We apply virtual camera to the last four (\emph{+vc}) and show the gains in \textcolor{dpurple}{purple}. $f$ denotes if a method handles variance in focal length.}
\label{tab:othermethods}
\vspace{-4mm}
\end{table}

\mypar{Ablations.}
Table~\ref{tab:ablations} ablates the features of \method on \dataset.
We report AP$_\threed$, performance at single IoU thresholds (0.25 and 0.50), and for near, medium and far objects. 
We further train and evaluate on domain-specific subsets, \datasetin (\APin) and \datasetout (\APout) to show how ablation trends hold for single domain scenarios.
From Table~\ref{tab:ablations} we draw a few standout observations:

\myparr{Virtual depth} is effective and improves AP$_\threed$ by $+6$\%, most noticable in the full \dataset which has the largest variances in camera intrinsics.
Enabled by virtual depth, scale augmentations during training increase AP$_\threed$ by $+3.1$\% on \dataset, $+2.7$\% on \datasetin and $+3.8$\% on \datasetout.
We find scale augmentation controlled \textit{without} virtual depth to be harmful by $-2.5\%$ (rows 6 - 7). 

\myparr{3D uncertainty} boosts performance by about $+6$\% on all \dataset subsets. 
Intuitively, it serves as a measure of confidence for the model's 3D predictions; removing it means that the model relies only on its 2D classification to score cuboids.
If uncertainty is used only to scale samples in Eq.~\ref{eq:loss}, but not at inference then AP$_\threed$ still improves but by $+1.6\%$. 

Table~\ref{tab:ablations} also shows that the entangled $\mathcal{L}_\text{3D}^\text{all}$ loss from Eq.~\ref{eq:loss} boosts AP$_\threed$ by $+3.1$\% while  disentangled losses contribute less ($+0.7$\%).
Replacing \emph{objectness} with \emph{IoUness} in the RPN head improves AP$_\threed$ on \dataset by $+1.1$\%. 

\mypar{Comparison to other methods.}
We compare \method to prior best approaches.
We choose representative state-of-the-art methods designed for the urban and indoor domains, and evaluate on our proposed \dataset benchmark and single-dataset benchmarks, KITTI and SUN RGB-D.

\myparr{Comparisons on \dataset.} 
Table~\ref{tab:othermethods} compares \method to M3D-RPN~\cite{brazil2019m3d}, a single-shot 3D anchor approach, FCOS3D~\cite{wang2021fcos3d} and its follow-up PGD~\cite{wang2022probabilistic}, which use a fully convolutional one-stage model, GUPNet~\cite{lu2021geometry}, which uses 2D-3D geometry and camera focal length to derive depth, SMOKE~\cite{liu2020smoke} which predicts 3D object center and offsets densely and ImVoxelNet~\cite{rukhovich2022imvoxelnet}, which uses intrinsics to unproject 2D image features to a 3D volume followed by 3D convolutions.
These methods originally experiment on urban domains, so we compare on \datasetout and the full \dataset.
We train each method using public code and report the best of multiple runs after tuning their hyper-parameters to ensure best performance.
We extend all but GUPNet and ImVoxelNet with a \emph{virtual camera} to handle varying intrinsics, denoted by \emph{+vc}.
GUPNet and ImVoxelNet dynamically use per-image intrinsics, which can naturally account for camera variation.
M3D-RPN and GUPNet do not support multi-node training, which makes scaling to \dataset difficult and are omitted.
For \datasetout, we report \APout on its test set and its subparts, AP$^\texttt{KIT}_\threed$ and AP$^\texttt{NU}_\threed$.

From Table~\ref{tab:othermethods} we observe that our \method outperforms competitive methods on both \datasetout and \dataset. 
The \emph{virtual camera} extensions result in performance gains of varying degree for all approaches, \eg $+4.2\%$ for PGD on \dataset, proving its impact as a general purpose feature for mixed dataset training for the task of 3D object detection.
FCOS3D, SMOKE, and ImVoxelNet~\cite{rukhovich2022imvoxelnet} struggle the most on \dataset, which is not surprising as they are typically tailored for dataset-specific model choices.

\begin{table}
\scalebox{0.85}{
\centering
\begin{tabular}{L{0.26\linewidth}|C{0.04\linewidth}|C{0.08\linewidth}C{0.08\linewidth}C{0.08\linewidth}|C{0.08\linewidth}C{0.08\linewidth}C{0.08\linewidth}}
&  & \multicolumn{3}{c|}{AP$_\threed^\textrm{70}$} & \multicolumn{3}{c}{AP$_\textrm{BEV}^\textrm{70}$} \\
Method & D & Easy & Med & Hard & Easy & Med & Hard \\
\hline
SMOKE~\cite{liu2020smoke} 				& \no & 14.03 & \phantom{0}9.76 & \phantom{0}7.84 & 20.83 & 14.49 & 12.75\\ 
ImVoxelNet~\cite{rukhovich2022imvoxelnet} 	& \no & 17.15 & 10.97 &  \phantom{0}9.15 & 25.19 & 16.37 & 13.58 \\
PGD~\cite{wang2022probabilistic}  			& \no & 19.05 & 11.76 &  \phantom{0}9.39 & 26.89 & 16.51 & 13.49 \\
GUPNet~\cite{lu2021geometry} 			& \no & 22.26 & \bf{15.02} & \bf{13.12} & 30.29 & 21.19 & 18.20 \\
\method 								& \no & \bf{23.59} & 15.01 & 12.56 & \bf{31.70} & \bf{21.20} & \bf{18.43} \\
\hline
MonoDTR~\cite{huang2022monodtr} 		& \yes & 21.99 & 15.39 & 12.73 & 28.59 & 20.38 & 17.14\\
DD3D~\cite{park2021dd3d}     				& \yes & 23.19 & 16.87 & 14.36 & 32.35 & 23.41 & 20.42 
\end{tabular} 
} 
\vspace{-3mm}
\caption{KITTI \href{https://www.cvlibs.net/datasets/kitti/eval_object.php?obj_benchmark=3d}{leaderboard} results on \emph{car} for image-based approaches. D denotes methods trained with extra depth supervision.}
\label{tab:kitti_test}
\vspace{-4mm}
\end{table}

\myparr{Comparisons on KITTI.} 
Table~\ref{tab:kitti_test} shows results on KITTI's test set using their server~\cite{kittiserver}. 
We train \method on KITTI only.
\method is a general purpose 3D object detector; its design is not tuned for the KITTI benchmark. 
Yet, it performs better or on par with recent best methods which are heavily tailored for KITTI and is only slightly worse to models trained with extra depth supervision.

\begin{table}
\centering
\scalebox{0.85}{
\begin{tabular}{L{.3\linewidth}|C{.3\linewidth}|C{.15\linewidth} C{.15\linewidth}}                                        
Method               & Trained on & $\text{\iou}_\threed$ & AP$_\threed$ \\
\hline
Total3D~\cite{Nie_2020_CVPR}    					& SUN RGB-D		& 23.3 	& - \\
ImVoxelNet~\cite{rukhovich2022imvoxelnet} 			& SUN RGB-D 		&  - 		& 30.6 \\
\methodshort 		                      					& SUN RGB-D		& 36.2 	& 34.7 \\
\methodshort 		                      					& \datasetin		& \bf{37.8} & \bf{35.4} \\
\end{tabular}
} 
\vspace{-3mm}
\caption{We compare to indoor models with common categories. For Total3D, we use oracle 2D detections to  report $\text{\iou}_\threed$ (3$^\textrm{rd}$ col). For ImVoxelNet, we report the detection metric of AP$_\threed$ (4$^\textrm{th}$ col). 
}
\label{tab:ours_vs_total3d}
\vspace{-4mm}
\end{table}

KITTI's AP$_\threed$ is fundamentally similar to ours but sets a single $\text{\iou}_\threed$ threshold at 0.70, which behaves similar to a nearly perfect 2D threshold of 0.94 (see the Appendix). 
To remedy this, our AP$_\threed$ is a mean over many thresholds, inspired by COCO~\cite{lin2014microsoft}.
In contrast, nuScenes proposes a mean AP based on 3D center distance but omits size and rotation, with settings generally tailored to the urban domain (\eg size and rotation variations are less extreme for cars).
We provide more analysis with this metric in the Appendix.

\myparr{Comparisons on SUN RGB-D.}
Table~\ref{tab:ours_vs_total3d} compares \method to Total3D~\cite{Nie_2020_CVPR} and ImVoxelNet~\cite{rukhovich2022imvoxelnet}, two state-of-the-art methods on SUN RGB-D on 10 common categories.
Total3D's public model \textit{requires} 2D object boxes and categories as input.
Hence, for a fair comparison, we use ground truth 2D detections as input to both our and their method, each trained using a ResNet34~\cite{he2016deep} backbone and report mean $\text{\iou}_\threed$ (3$^\textrm{rd}$ col).
We compare to ImVoxelNet in the full 3D object detection setting and report AP$_\threed$ (4$^\textrm{th}$ col).
We compare \method from two respects, first trained on SUN RGB-D identical to the baselines, then trained on \datasetin which subsumes SUN RGB-D.
Our model outperforms Total3D by $+12.9\%$ and ImVoxelNet by $+4.1\%$ when trained on the same training set and increases the performance gap when trained on the larger \datasetin.

\myparr{Zero-shot Performance.}
One commonly sought after property of detectors is zero-shot generalization to other datasets.
\method outperforms all other methods at zero-shot.
We find our model trained on KITTI achieves 12.7\% AP$_\threed$ on nuScenes; the second best is M3D-RPN\emph{+vc} with 10.7\%.
Conversely, when training on nuScenes, our model achieves 20.2\% on KITTI; the second best is GUPNet with 17.3\%.

\subsection{The Impact of the \dataset Benchmark}

Sec.~\ref{sec:ablations} analyzes the performance of \method for the task of 3D object detection.
Now, we turn to \dataset and its impact as a large-scale benchmark.
We show two use cases of \dataset: (a) a universal 3D dataset which integrates smaller ones, and (b) a pre-training dataset.

\mypar{\dataset as a universal dataset.}
We treat \dataset as a dataset which integrates smaller single ones and show its impact on each one.
We train \method on \dataset and compare to single-dataset training in Table~\ref{tab:cross_dataset}, for the indoor (left) and urban (right) domain.
AP$_\threed$ is reported on the category intersection (10 for indoor, 3 for outdoor) to ensure comparability.
Training on \dataset and its domain-specific subsets, \datasetin and \datasetout, results in higher performance compared to single-dataset training, signifying that our large \dataset generalizes better and should be preferred over single dataset training.
ARKit sees a $+2.6\%$ boost and KITTI $+5.3$\%.
Except for Hypersim, the domain-specific subsets tend to perform better on their domain, which is not surprising given their distinct properties (Fig.~\ref{fig:dataset_stats}).

\mypar{\dataset as a pre-training dataset.}
Next, we demonstrate the utility of \dataset for pre-training. 
In this setting, an unseen dataset is used for finetuning from a model pre-trained on \dataset.
The motivation is to determine how a large-scale 3D dataset could accelerate low-shot learning with minimum need for costly 3D annotations on a new dataset.
We choose SUN RGB-D and KITTI as our \textit{unseen} datasets given their popularity and small size.
We pre-train \method on \dataset-, which removes them from \dataset, then finetune the models using a \% of their training data. 
The curves in Fig.~\ref{fig:percentoftraining} show the model quickly gains its upper-bound performance at a small fraction of the training data, when pre-trained on \dataset- vs. ImageNet, without any few-shot training tricks. A model finetuned on only 5\% of its target can achieve $>$70\% of the upper-bound performance. 

\begin{table}[t]
\begin{minipage}{0.49\linewidth}
\raggedright
\scalebox{0.84}{
\begin{tabular}{L{.4\linewidth}|C{.15\linewidth}C{.15\linewidth}C{.15\linewidth}}
                                     
Trained on  & AP$^\texttt{HYP}_\threed$	& AP$^\texttt{SUN}_\threed$  	& AP$^\texttt{AR}_\threed$ \\
\hline
Hypersim   &   15.2 &   $~~$9.5       & $~~$7.5       \\
SUN        & $~~$5.8 	&  34.7  	& 13.1    \\
ARKit       &  $~~$5.9 &  14.2    	& 38.6  	 \\
\dataset     &  \bf{19.0}  & 32.6 & 38.2  	\\
\datasetin     & 17.8    	& \bf{35.4} & \bf{41.2}   \\
\end{tabular}
} 
\end{minipage}
\begin{minipage}{0.49\linewidth}
\raggedleft
\scalebox{0.84}{
\begin{tabular}{L{.45\linewidth}|C{.15\linewidth}C{.15\linewidth}}
Trained on 		& AP$^\texttt{KIT}_\threed$  	& AP$^\texttt{NU}_\threed$ 	 \\
\hline
KITTI         		&  37.1  	& 12.7  	 \\
nuScenes       		&  20.2    	& 38.6     \\
\dataset     		&  37.8    	& 35.8     \\
\datasetout     		&  \bf{42.4}    	& \bf{39.0}     \\
\multicolumn{3}{l}{} \\
\end{tabular}
} 
\end{minipage}
\vspace{-3mm}
\caption{Cross-dataset performance on \textit{intersecting} categories (10 for indoor, 3 for outdoor) for comparable cross-evaluations.}
\label{tab:cross_dataset}
\vspace{2mm}
\centering
\includegraphics[width=0.84\linewidth]{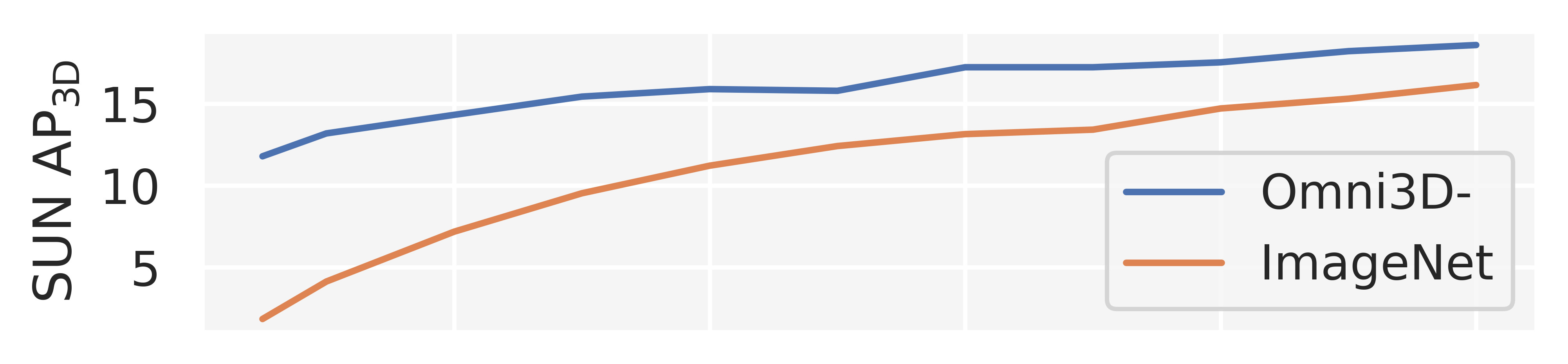}
\includegraphics[width=0.84\linewidth]{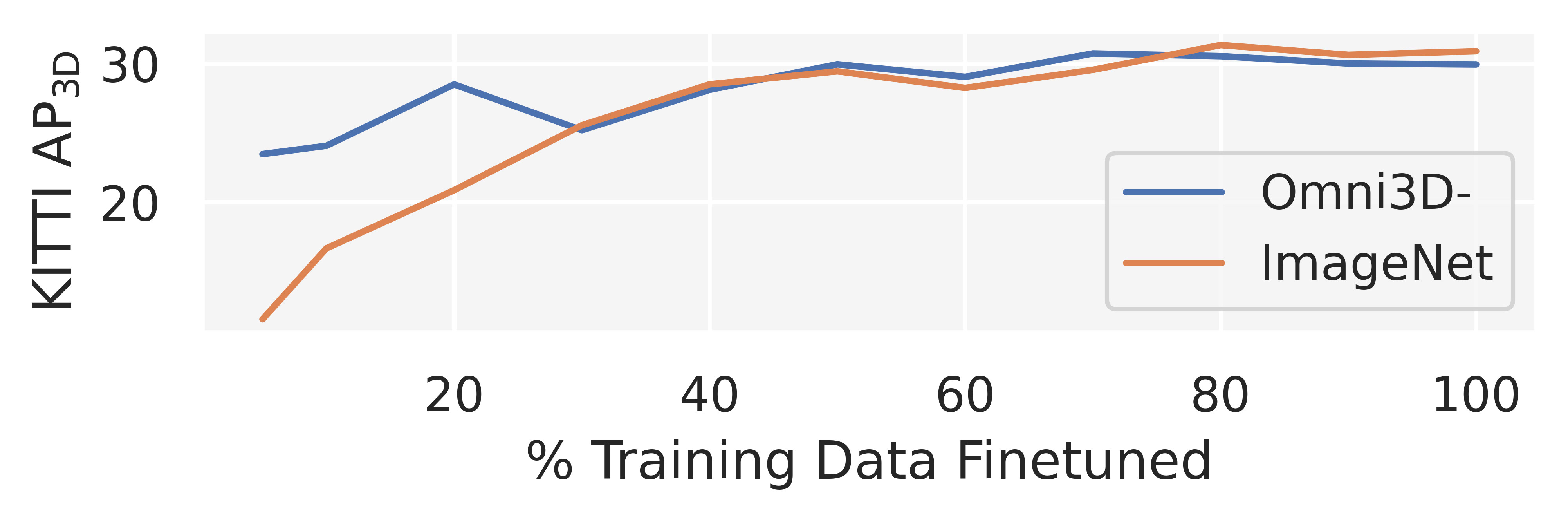} 
\vspace{-3mm}
\captionof{figure}{Pre-training on \dataset- vs. ImageNet.}
\label{fig:percentoftraining}
\vspace{-5mm}
\end{table}

\begin{figure*}[t]
  \setlength{\linewidth}{\textwidth}
  \setlength{\hsize}{\textwidth}
  \centering
  { \footnotesize \hspace{-1mm} \dataset  \hspace{2.0mm} \datasetin  \hspace{0.2mm} \datasetout  \hspace{0.2mm}  SUN RGB-D  \hspace{2.0mm} ARKit  \hspace{3.5mm} Hypersim  \hspace{3.2mm} Objectron  \hspace{4.6mm}  KITTI  \hspace{4.5mm} nuScenes \hspace{5.8mm} COCO \hspace{6.5mm} LVIS}
  \includegraphics[width=1.0\linewidth]{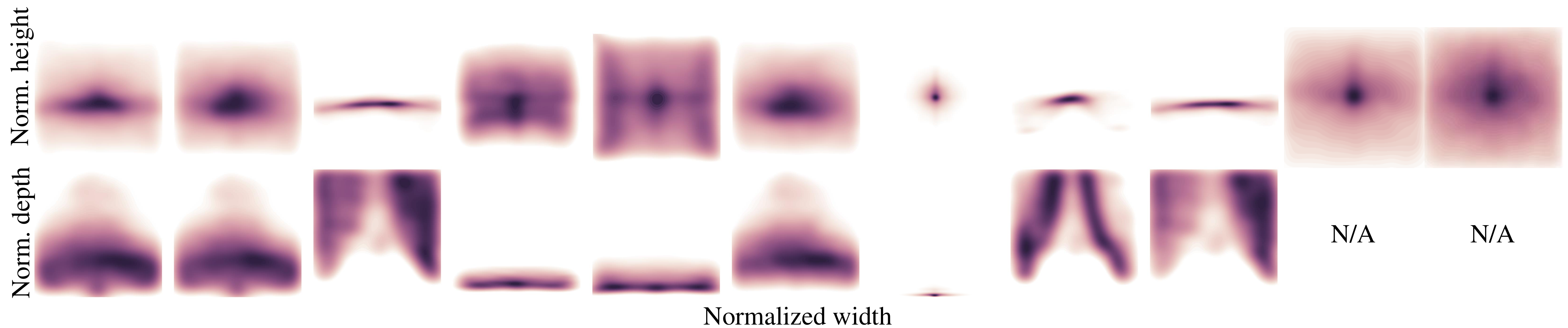}
  \vspace{-8mm}
  \caption{
  We show the distribution of object centers in normalized image coordinates projected onto the XY-plane (top) and normalized depth projected onto the topview XZ-plane (bottom) across each subset of \dataset, individual 3D datasets SUN RGB-D~\cite{song2015sun}, ARKit~\cite{dehghan2021arkitscenes}, Hypersim~\cite{hypersim}, Objectron~\cite{objectron2021}, KITTI~\cite{Geiger2012CVPR}, nuScenes~\cite{caesar2020nuscenes}, and large-scale 2D datasets COCO~\cite{lin2014microsoft} and LVIS~\cite{gupta2019lvis}.
}\vspace{-1mm}
  \label{fig:dataset_stats_suppl}
  \end{figure*}  

\vspace{-1mm}
\section{Conclusion}
\label{sec:conclusion}
\vspace{-1mm}
We propose a large and diverse 3D object detection benchmark, \dataset, and a general purpose 3D object detector, \method.
Models and data are publicly released. 
Our extensive analysis (Table~\ref{tab:ablations}, Fig.~\ref{fig:qualitative}, Appendix) show the strength of general 3D object detection as well as the limitations of our method, \eg localizing far away objects, and uncommon object types or contexts.
Moreover, for accurate real-world 3D predictions, our method is limited by the assumption of known camera intrinsics, which may be minimized in future work using self-calibration techniques. 
\setcounter{section}{0}
\renewcommand{\thesection}{A\arabic{section}}
\paragraph{\noindent{\textbf{\Large{Appendix}}}}
\section{Dataset Details}
In this section, we provide details related to the creation of \dataset, its sources, coordinate systems, and statistics. 

\begin{figure*}[t]
\centering
\includegraphics[width=0.99\linewidth]{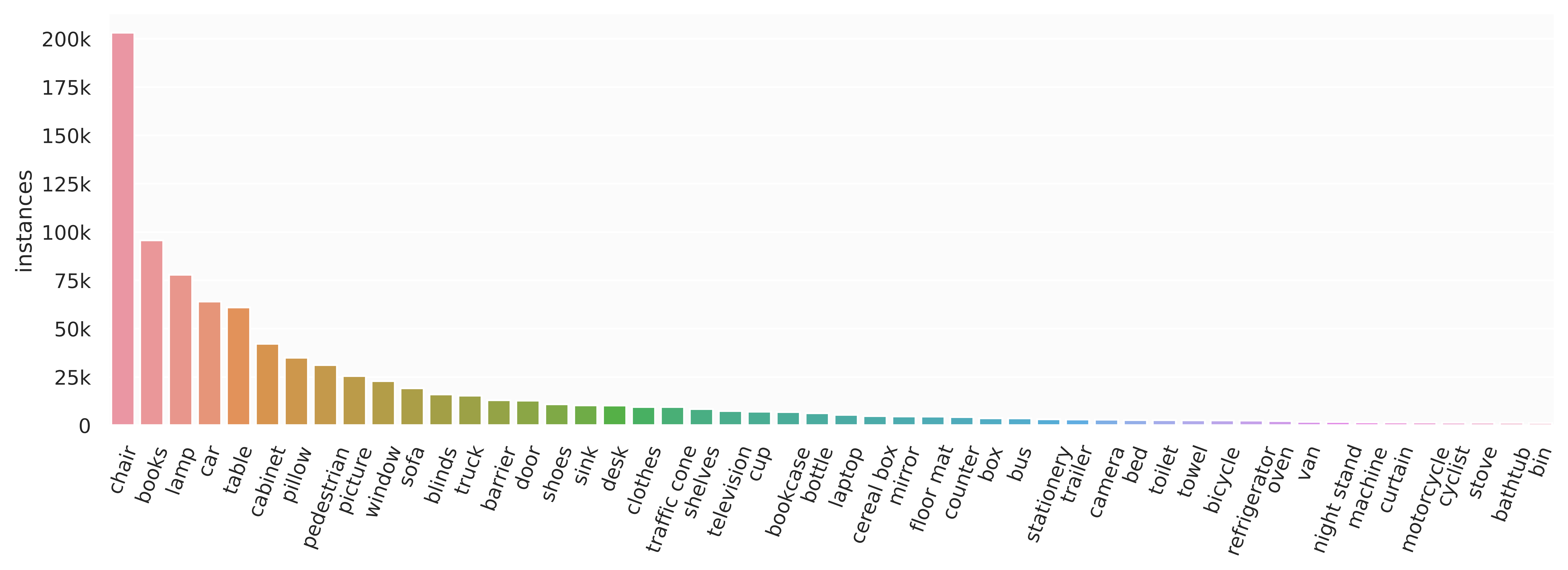}
\caption{Number of instances for the 50 categories in \dataset.}
\label{fig:category_counts}
\end{figure*}

\mypar{Sources.} For each individual dataset used in \dataset, we use their official train, val, and test set for any datasets which have all annotations released. 
If no public test set is available then we use their validation set \textit{as} our test set. 
Whenever necessary we further split the remaining training set in 10:1 ratio by sequences in order form train and val sets.
The resultant image make up of \dataset is detailed in Table~\ref{tab:dataset_stats}.
To make the benchmark coherent we merged semantic categories across the sources, \eg, there are 26 variants of chair
including ‘chair’, ‘chairs’, ‘recliner’, ‘rocking chair’, \etc
We show the category instance counts in Figure~\ref{fig:category_counts}.

\mypar{Coordinate system.} We define our unified 3D coordinate system for all labels with the camera center being the origin and  +x facing right, +y facing down, +z inward~\cite{Geiger2012CVPR}.
Object pose is relative to an initial object with its bottom-face normal aligned to +y and its front-face aligned to +x (\eg upright and facing to the right). 
All images have known camera calibration matrices with input resolutions varying from $370$ to $1920$ and diverse focal lengths from $518$ to $1708$ in pixels. 
Each object label contains a category label, a 2D bounding box, a 3D centroid in camera space meters, a $3\times3$ matrix defining the object to camera rotation, and the physical dimensions (width, height, length) in meters.

\begin{figure*}[t]
\centering
\includegraphics[width=0.99\linewidth]{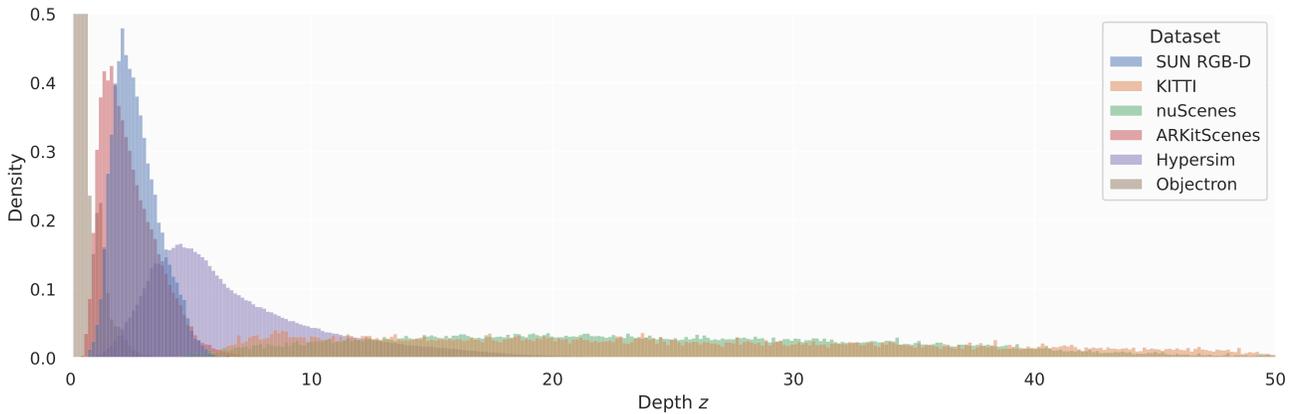}
\caption{Normalized depth distribution per dataset in \dataset. We slightly limit the depth and density ranges for a better visualization.}
\label{fig:depth_distribution}
\end{figure*}

\mypar{Spatial Statistics.}
Following Section 3 of the main paper, we provide the spatial statistics for the individual data sources of SUN RGB-D~\cite{song2015sun}, ARKit~\cite{dehghan2021arkitscenes}, Hypersim~\cite{hypersim}, Objectron~\cite{objectron2021}, KITTI~\cite{Geiger2012CVPR}, nuScenes~\cite{caesar2020nuscenes}, as well as \datasetin and \datasetout in Figure~\ref{fig:dataset_stats_suppl}.
As we mention in the main paper, we observe that the indoor domain data, with the exception of Hypersim, have bias for close objects. 
Moreover, outdoor data tend to be spatially biased with projected centroids along diagonal ground planes while indoor is more central.
We provide a more detailed view of density-normalized depth distributions for each dataset in Figure~\ref{fig:depth_distribution}.

\begin{table}
\scalebox{0.85}{
\centering
\begin{tabular}{L{0.30\linewidth}|C{0.03\linewidth}C{0.06\linewidth}|C{0.12\linewidth}C{0.12\linewidth}C{0.10\linewidth}C{0.11\linewidth}}
Method & $|C|$ & $|C^*|$ & Total & Train & Val & Test \\
\hline
KITTI~\cite{Geiger2012CVPR} & 8 & \phantom{0}5 & \phantom{00}7,481 & \phantom{00}3,321 & \phantom{00}391 & \phantom{0}3,769 \\
SUN RGB-D~\cite{song2015sun} & 83 & 38 & \phantom{0}10,335 & \phantom{00}4,929 & \phantom{00}356 & \phantom{0}5,050 \\
nuScenes~\cite{caesar2020nuscenes} & 9 & \phantom{0}9 & \phantom{0}34,149 & \phantom{0}26,215 & \phantom{0}1,915 & \phantom{0}6,019 \\
Objectron~\cite{objectron2021} & 9 & \phantom{0}9 & \phantom{0}46,644 & \phantom{0}33,519 & \phantom{0}3,811 & \phantom{0}9,314 \\ 
ARKitScenes~\cite{dehghan2021arkitscenes} & 15 & 14  & \phantom{0}60,924  & \phantom{0}48,046  & \phantom{0}5,268  & \phantom{0}7,610 \\
Hypersim~\cite{hypersim} & 32 & 29 & \phantom{0}74,619  & \phantom{0}59,543  & \phantom{0}7,386  & \phantom{0}7,690 \\ \hline
\datasetout  & 14 & 11 & \phantom{0}41,630  & \phantom{0}29,536  & \phantom{0}2,306  & \phantom{0}9,788 \\
\datasetin  & 84 & 38 & 145,878  & 112,518  & 13,010  & 20,350 \\
\dataset  & 98 & 50 & 234,152  & 175,573  & 19,127  & 39,452 \\
\end{tabular} 
} 
\vspace{-1mm}
\caption{
We detail the statistics for each dataset split. We report the total number of categories $|C|$, and the number of categories $|C^*|$ used in our paper and our AP$_\threed$ metrics. $C^*$ contains all categories from $C$ with at least 1000 positive instances. Finally, we report the total number of images split into train/val/test.
}
\label{tab:dataset_stats}
\vspace{2mm}
\end{table}

\begin{figure*}[t]
\centering
\includegraphics[width=0.99\textwidth]{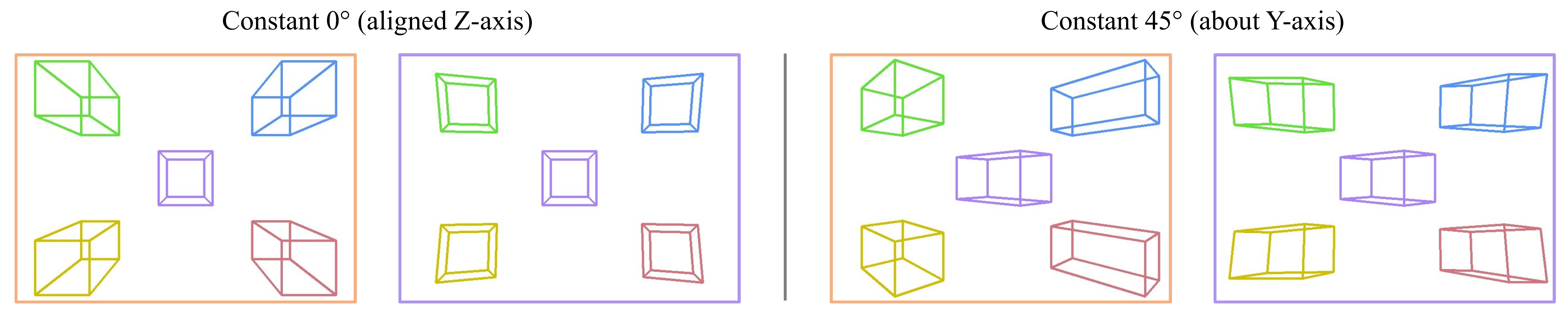}
\vspace{-2mm}
\caption{We show \textcolor{myorange}{egocentric} and \textcolor{mypurple}{allocentric} representations under constant rotations. Despite being identical rotations, the egocentric representation appears visually different as its spatial location changes, unlike allocentric which is consistent with location changes. }
\label{fig:allocentric}
\vspace{-4mm}
\end{figure*}

\section{Model Details}

In this section, we provide more details for \method pertaining to its 3D bounding box allocentric rotation (Sec. 4.1)  and the derivation of virtual depth (Sec. 4.2). 

\subsection{3D Box Rotation}

Our 3D bounding box object rotation is predicted in the form of a 6D continuous parameter, which is shown in~\cite{zhou2019continuity} to be better suited for neural networks to regress compared to other forms of rotation.
Let our predicted rotation $\bf{p}$ be split into two directional vectors $\bf{p}_{1},\bf{p}_{2} \in \mathbb{R}^3$ and $\bf{r}_{1},\bf{r}_{2},\bf{r}_{3}  \in \mathbb{R}^3$ be the columns of a $3\times3$ rotational matrix $\bf{R_a}$. 
Then $\bf{p}$ is mapped to $\bf{R_a}$ via
\begin{align}
\bf{r}_1 &= \text{norm}(\bf{p}_1) \\
\bf{r}_2 &= \text{norm}(\bf{p}_2 - (\bf{r}_1 \cdot \bf{p}_2)\bf{r}_1) \\
\bf{r}_3 & = \bf{r}_1 \times \bf{r}_2
\end{align} 

where $(\cdot,\times)$ denote dot and cross product respectively.   

$\bf{R}_a$ is estimated in \textit{allocentric} form similar to~\cite{kundu20183d}.
Let $f_x, f_y, p_x, p_y$ be the known camera intrinsics, $u, v$ the predicted 2D projected center as in Section 4.1, and $a =[0, 0, 1]$ be the camera's principal axis.
Then $o = \text{norm}([\frac{u-p_x}{f_x},~\frac{v-p_y}{f_y}, 1])$ is a ray pointing from the camera to $u, v$ with angle $\alpha = \text{acos}(o)$.
Using standard practices of axis angle representations with an axis denoted as $o \times a$ and angle $\alpha$, we compute a matrix $\mathbf{M}\in \mathbb{R}^{3\times3}$, which helps form the final \textit{egocentric} rotation matrix $\mathbf{R} = \bf{M}\cdot\bf{R}_a$.

We provide examples of 3D bounding boxes at constant egocentric or allocentric rotations in Figure~\ref{fig:allocentric}. 
The allocentric rotation is more aligned to the visual 2D evidence, whereas egocentric rotation entangles relative position into the prediction. 
In other words, identical egocentric rotations may look very different when viewed from varying spatial locations, which is \textit{not} true for allocentric.

\subsection{Virtual Depth}

In Section 4.2, we propose a virtual depth transformation in order to help \method handle varying input image resolutions and camera intrinsics. 
In our experiments, we show that virtual depth also helps other competing approaches, proving its effectiveness as a general purpose feature.
The motivation of estimating a \textit{virtual} depth instead of \textit{metric} depth is to keep the effective image size and focal length consistent in an invariant camera space. 
Doing so enables two camera systems with nearly the same visual evidence of an object to transform into the same virtual depth as shown in Figure 4 of the main paper. 
Next we provide the proof for the conversion between virtual and metric depth. 

\noindent \emph{Proof:}
Assume a 3D point $(X,Y,Z)$ projected to $(x,y)$ on an image with height $H$ and focal length $f$.\linebreak
The virtual 3D point $(X, Y, Z_v)$ is projected to $(x_v,y_v)$ on the virtual image.
The 2D points $(x,y)$ and $(x_v, y_v)$ correspond to the same pixel location in both the original and virtual image.
In other words, $y_v = y \cdot \frac{H_v}{H}$.
Recall the formula for projection as $y = f\cdot\frac{Y}{Z} + p_y$ and $y_v = f_v\cdot\frac{Y}{Z_v} + p_{y_v}$, where $p_y$ is the principal point and $p_{y_v} = p_y \cdot \frac{H_v}{H}$.
By substitution $f_v \cdot \frac{Y}{Z_v} = f \cdot \frac{Y}{Z} \frac{H_v}{H} \Rightarrow Z_v = Z \cdot \frac{f_v}{f} \frac{H}{H_v}$.

\subsection{Training Details and Efficiency}

When training on subsets smaller than \dataset, we adjust the learning rate, batch size, and number of iterations linearly until we can train for 128 epochs between 96k to 116k iterations.
\method trains on V100 GPUs between 14 and 26 hours depending on the subset configuration when scaled to multi-node distributed training, and while training uses approximately 1.6 GB memory per image. 
\textit{Inference} on a \method model processes image from KITTI~\cite{Geiger2012CVPR} with a wide input resolution of 512$\times$1696 at 52ms/image on average while taking up 1.3 GB memory on a Quadro GP100 GPU.
Computed with an identical environment, our model efficiency is favorable to M3D-RPN~\cite{brazil2019m3d} and GUPNet~\cite{lu2021geometry} which infer from KITTI images at 191ms and 66ms on average, respectively.

\section{Evaluation}

In this section we give additional context and justification for our chosen thresholds $\tau$ which define the ranges of 3D IoU that AP$_\threed$ is averaged over. 
We further provide details on our implementation of 3D IoU. 

\begin{figure*}[t]
\centering
\includegraphics[width=0.99\linewidth]{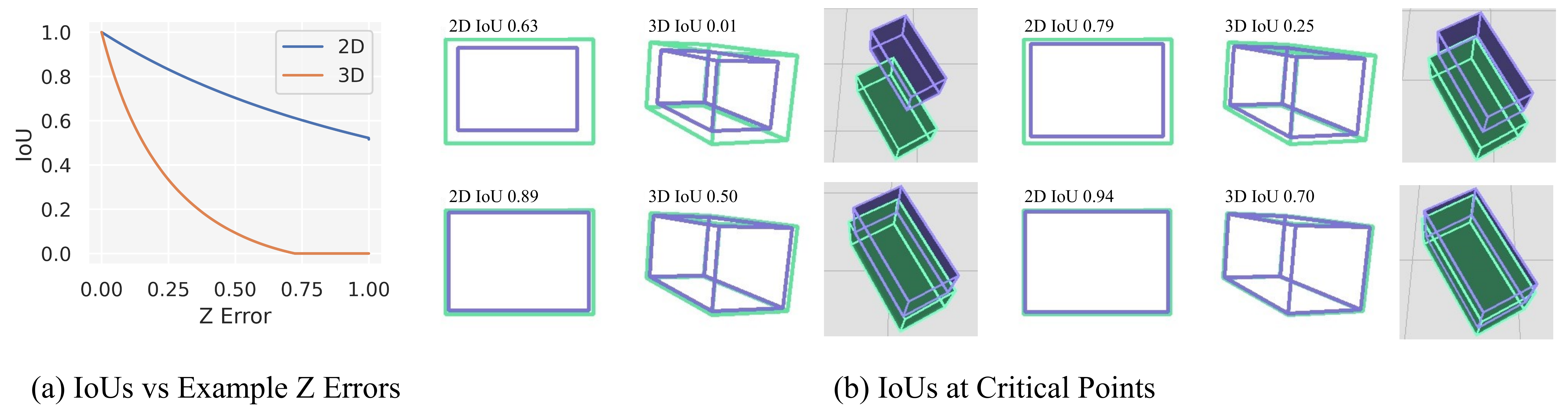}
\caption{We slowly translate a 3D box (\textcolor{mypurple}{purple}) backwards relative to its ground-truth (\textcolor{mygreen}{green}), hence simulating error in $z$ up to 1 unit length ($\bar{l}^\text{gt}_\threed = 1$). 
We plot 2D and 3D IoU vs $z$ error in (a) and visualize selected critical points showing their 2D projections, 3D front view and a novel top view.
Unsurprisingly, we find the drop in IoU$_\threed$ is much steeper than 2D. This effect highlights both the challenge of 3D object detection and helps justify the relaxed $\tau$ thresholds used for AP$_\threed$ in Section $5$ of the main paper.}
\label{fig:ious}
\end{figure*}

\subsection{Thresholds $\tau$ for AP$_\threed$}
We highlight the relationship between 2D and 3D IoU in Figure~\ref{fig:ious}.
We do so by contriving a general example between a ground truth and a predicted box, both of which share rotation and rectangular cuboid dimensions ($0.5\times 0.5 \times 1.0$).
We then translate the 3D box along the $z$-axis up to 1 meter (its unit length), simulating small to large errors in depth estimation. 
As shown in the left of Figure~\ref{fig:ious}, the 3D IoU drops off significantly quicker than 2D IoU does between the projected 2D bounding boxes. 
As visualized in right of Figure~\ref{fig:ious}, a moderate score of 0.63 IoU$_\twod$ may result in a low $0.01$ IoU$_\threed$. 
Despite visually appearing to be well localized in the front view, the top view helps reveal the error.  
Since depth is a key error mode for 3D, we find the relaxed settings of $\tau$ compared to 2D (Sec. 5) to be reasonable.

\subsection{$\text{\iou}_\threed$ Details}

We implement a fast $\text{\iou}_\threed$. We provide more details for our algorithm here. 
Our algorithm starts from the simple observation that the intersection of two oriented 3D boxes, $b_1$ and $b_2$, is a convex polyhedron with $n > 2$ comprised of connected planar units. In 3D, these planar units are 3D triangular faces. Critically, each planar unit belongs strictly to either $b_1$ or $b_2$. Our algorithm finds these units by iterating through the sides of each box, as described in Algorithm~\ref{alg:iou3d}.

\begin{algorithm}[t]
\setstretch{1.35}
\KwData{Two 3D boxes $b_1$ and $b_2$}  
\KwResult{Intersecting shape $S = []$}
Step 1: For each 3D triangular face $e \in b_1$ we check wether $e$ falls inside $b_2$ \\
Step 2: If $e$ is not inside, then we discard it \\
Step 3: If $e$ is inside, then $S = S + [e]$. If $e$ is partially inside, then the part of $e$ inside $b_2$, call it $\hat{e}$, is added to $S$, $S = S + [\hat{e}]$ \\
Step 4: We repeat steps 1 - 3 for $b_2$ \\
Step 5: We check and remove duplicates in $S$ (in the case of coplanar sides in $b_1$ and $b_2$)\\
Step 6: We compute the volume of $S$, which is guaranteed to be convex
\caption{A high-level overview of our fast and exact $\text{\iou}_\threed$ algorithm.}
\label{alg:iou3d}
\end{algorithm}

\section{nuScenes Performance}

In the main paper, we compare \method to competing methods with the popular IoU-based AP$_\threed$ metric, which is commonly used and suitable for general purpose 3D object detection.
Here, we additionally compare our \method to best performing methods with the nuScenes evaluation metric, which is designed for the urban domain.
The released nuScenes dataset has 6 cameras in total and reports custom designed metrics of mAP and NDS as detailed in~\cite{caesar2020nuscenes}.
Predictions from the 6 input views (each from a different camera) are fused to produce a single prediction.
In contrast, our benchmark uses 1 camera (front) and therefore does not require or involve any post-processing fusion nor any dataset-specific predictions (e.g velocity and attributes). 
Moreover, as discussed in Section 5.1, the mAP score used in nuScenes is based on distances of 3D object centers which ignores errors in rotation and object dimensions. 
This is suited for urban domains as cars tend to vary less in size and orientation and is partially addressed in the NDS metric where true-positive (TP) metrics are factored in as an average over mAP and TP (Eq. 3 of ~\cite{caesar2020nuscenes}).

\begin{table}
\scalebox{0.85}{
\centering
\begin{tabular}{L{0.23\linewidth}|C{0.06\linewidth}|C{0.08\linewidth}C{0.08\linewidth}C{0.08\linewidth}C{0.08\linewidth}C{0.1\linewidth}|C{0.08\linewidth}}
& &  \multicolumn{5}{c|}{\textbf{nuScenes Front Camera Only}}\\
Method & TTA & mAP & mATE & mASE & mAOE & NDS & AP$_\threed$\\ \hline
FCOS3D~\cite{wang2021fcos3d} & \yes & 35.3 & 0.777  & 0.231 & 0.400  & 44.2 & 27.9 \\
PGD~\cite{wang2022probabilistic}  & \yes & 39.0 & 0.675 & 0.236 & 0.399 & 47.6  & 32.3 \\
\method & \no  & 32.6 & 0.671 & 0.289 & 1.000 & 33.6 & 33.0 \\
\end{tabular} 
} 
\vspace{-3mm}
\caption{
We compare \method to competing methods, FCOS3D and PGD, on the nuScenes metrics on the single front camera setting, and without velocity or attribute computations factored into the NDS metric. 
The nuScenes metric uses a center-distance criteria at thresholds of \{0.5, 1, 2, 4\} meters, ignoring object size and rotation, whereas our AP$_\threed$ metric (last col.) uses IoU$_\threed$.
\method predicts relative orientation to maximize IoU$_\threed$ rather than absolute, and therefore receives a high mAOE. Note that FCOS3D and PGD use test-time augmentations (TTA); we don't.
}
\label{tab:nuscenes}
\vspace{-4mm}
\end{table}

We compare with the current best performing methods on nuScenes, FCOS3D~\cite{wang2021fcos3d} and PGD~\cite{wang2022probabilistic}.
We take their best models including fine-tuning and test-time augmentations (TTA), then modify the nuScenes evaluation metric in the following three ways:

\begin{enumerate}
\item{ We evaluate on the single front camera setting.}
\item{ We merge the construction vehicle and truck categories, as in the pre-processing of \dataset.}
\item{ We drop the velocity and attribute true-positive metrics from being included in the NDS metric. Following~\cite{caesar2020nuscenes} our TP $=$ \{mATE, mASE, mAOE\} resulting in a metric of $\text{NDS} = \frac{1}{6} ~(3~\text{mAP} + \sum\limits_{\text{mTP}\in \text{TP}}{1 - \text{min}(1, \text{mTP})})$}
\end{enumerate}

Table~\ref{tab:nuscenes} reports the performance of each nuScenes metric for \method, FCOS3D, and PGD along with our AP$_\threed$. 
Since these methods are evaluated on the front cameras, no late-fusion of the full camera array is performed and side/back camera ground truths are not evaluated with.
\method performs competitively but slightly worse on the mAP center-distance metric compared to FCOS3D and PGD.
This is not surprising as the FCOS3D and PGD models were tuned for the nuScenes benchmark and metric and additionally use test time augmentations, while we don't.

\begin{table*}[t]
\scalebox{0.95}{
\begin{tabular}{L{.14\textwidth}|C{.12\textwidth}|C{.03\textwidth}C{.03\textwidth}C{.03\textwidth}C{.05\textwidth}C{.03\textwidth}C{.05\textwidth}C{.05\textwidth}C{.04\textwidth}C{.03\textwidth}C{.07\textwidth}|C{.03\textwidth}}                                        
Method               & Trained on 					& table & bed & sofa & bathtub & sink & shelves & cabinet & fridge & chair & television & avg.  \\
\hline
ImVoxelNet~\cite{rukhovich2022imvoxelnet}    & SUN RGB-D	&   39.5  &  68.8  &  48.9     &  33.9  &  18.7    & \phantom{0}2.4 & 13.2 & 17.0 &  55.5   & \phantom{0}8.4 & 30.6\\
\methodshort 		                      & SUN RGB-D	&   39.2  &  65.7  &  58.1     &  49.0  &  32.5    & \phantom{0}4.3 & 16.2 & 25.2 &  54.5   & \phantom{0}2.7 & 34.7\\
\methodshort 		                      &	\datasetin		&  38.3  &  66.5  &  60.3     &  51.8  &  30.8    &   \phantom{0}3.2  &  13.3 &  30.3   & 56.1   &  \phantom{0}3.6 & \bf{35.4}\\
\end{tabular}
} 
\vspace{-2mm}
\caption{Comparison to ImVoxelNet~\cite{rukhovich2022imvoxelnet}  on SUN RGB-D test. We use the full 3D object detection setting to report \bf{AP$_\threed$}.}
\label{tab:fullimvoxelnet}
\vspace{-1mm}
\end{table*}

\begin{table*}[t]
\scalebox{0.95}{
\begin{tabular}{L{.14\textwidth}|C{.12\textwidth}|C{.03\textwidth}C{.03\textwidth}C{.03\textwidth}C{.05\textwidth}C{.03\textwidth}C{.05\textwidth}C{.05\textwidth}C{.04\textwidth}C{.03\textwidth}C{.07\textwidth}|C{.03\textwidth}}                                        
Method               & Trained on 					& table & bed & sofa & bathtub & sink & shelves & cabinet & fridge & chair & toilet & avg.  \\
\hline
Total3D~\cite{Nie_2020_CVPR}    & SUN RGB-D	&  27.7  &  33.6    &  30.1   & 28.5   & 18.8     & 10.1  & 13.1 & 19.1    & 24.2  & 28.1 &  23.3\\
\methodshort 		                      & SUN RGB-D	&  39.2  &  49.5  &  46.0     &  32.2  &  31.9    & 16.2 & 26.5 & 34.7    & 39.9   & 45.7 & 36.2\\
\methodshort 		                      &	\datasetin		& 40.7  &  50.1  &  50.0     &  33.8  &  31.8    & 18.2  & 29.0 &  34.6   & 41.6   & 48.2 & \bf{37.8}\\
\end{tabular}
} 
\vspace{-2mm}
\caption{Comparison to Total3D~\cite{Nie_2020_CVPR} on SUN RGB-D test. We use oracle 2D detections fairly for all methods and report \bf{\texttt{IoU}$_\threed$}.}
\label{tab:fulltotal3d}
\vspace{-4mm}
\end{table*}

\section{Full Category Performance on \dataset}
We train \method on the full $98$ categories within \dataset, in contrast to the main paper which uses the $50$ most frequent categories with more than $1k$ positive instances.
As expected, the AP$_\threed$ performance decreases when evaluating on the full categories from $23.3$ to $14.1$, and similarly for AP$_\twod$ from $27.6$ to $17.3$. 
At the long-tails, we expect 2D and 3D object recognition will suffer since fewer positive examples are available for learning.
We expect techniques related to few-shot recognition could be impactful and lend to a new avenue for exploration with \dataset.

\section{Per-category SUN RGB-D Performance}

We show per-category performance on AP$_\threed$ for \method and ImVoxelNet~\cite{rukhovich2022imvoxelnet}'s publicly released indoor model in Table~\ref{tab:fullimvoxelnet}.
We show the 10 common categories which intersect all indoor datasets as used in Table 4-5 in the main paper for fair comparisons when training categories differ.

Similarly, Table~\ref{tab:fulltotal3d} shows detailed per-category $\text{\iou}_\threed$ performance for \method and Total3D~\cite{Nie_2020_CVPR} on 10 common categories, a summary of which was presented in Table 4 in the main paper.
Note that \emph{television}, in our 10 intersecting categories, is not detected by Total3D thus we replace it by \emph{toilet} which is the next most common category.

\section{Regarding the Public Models Used for \dataset Comparisons}

We use publicly released code for M3D-RPN~\cite{brazil2019m3d}, GUPNet~\cite{lu2021geometry}, SMOKE~\cite{liu2020smoke}, FCOS3D~\cite{wang2021fcos3d}, PGD~\cite{wang2022probabilistic}, and ImVoxelNet~\cite{rukhovich2022imvoxelnet} and Total3D~\cite{Nie_2020_CVPR} in Section 5.  Most of these models are implemented in the mmdetection3d~\cite{mmdet3d2020} open-source repository, which critically supports many features for dataset scaling such as distributed scaling and strategic data sampling. 

Most of the above methods tailor their configuration hyper-parameters in a handful of ways specifically for each dataset they were originally designed for. 
Since our experiments explicitly run on mixed datasets which have diverse resolutions, aspect ratios, depth distributions, \etc, we opted to run each method on a variety of settings.  
We did our best to run each method on multiple sensible hyper-parameters settings to give each method the most fair chance.
We focused primarily on the settings which are impacted by input resolution and depth distributions, as these are seemingly the biggest distinctions between datasets.
When applicable, we implemented virtual depth with a custom design for each method depending on its code structure (see main paper).
 
Although we expect that better recipes can be found for each method, \textbf{we ran more than 100 experiments} and are using 17 of these runs \textbf{selected based on best performance} to report in the paper. 

\section{Qualitative Examples}
\begin{figure*}[t]
\centering
\includegraphics[width=0.99\linewidth]{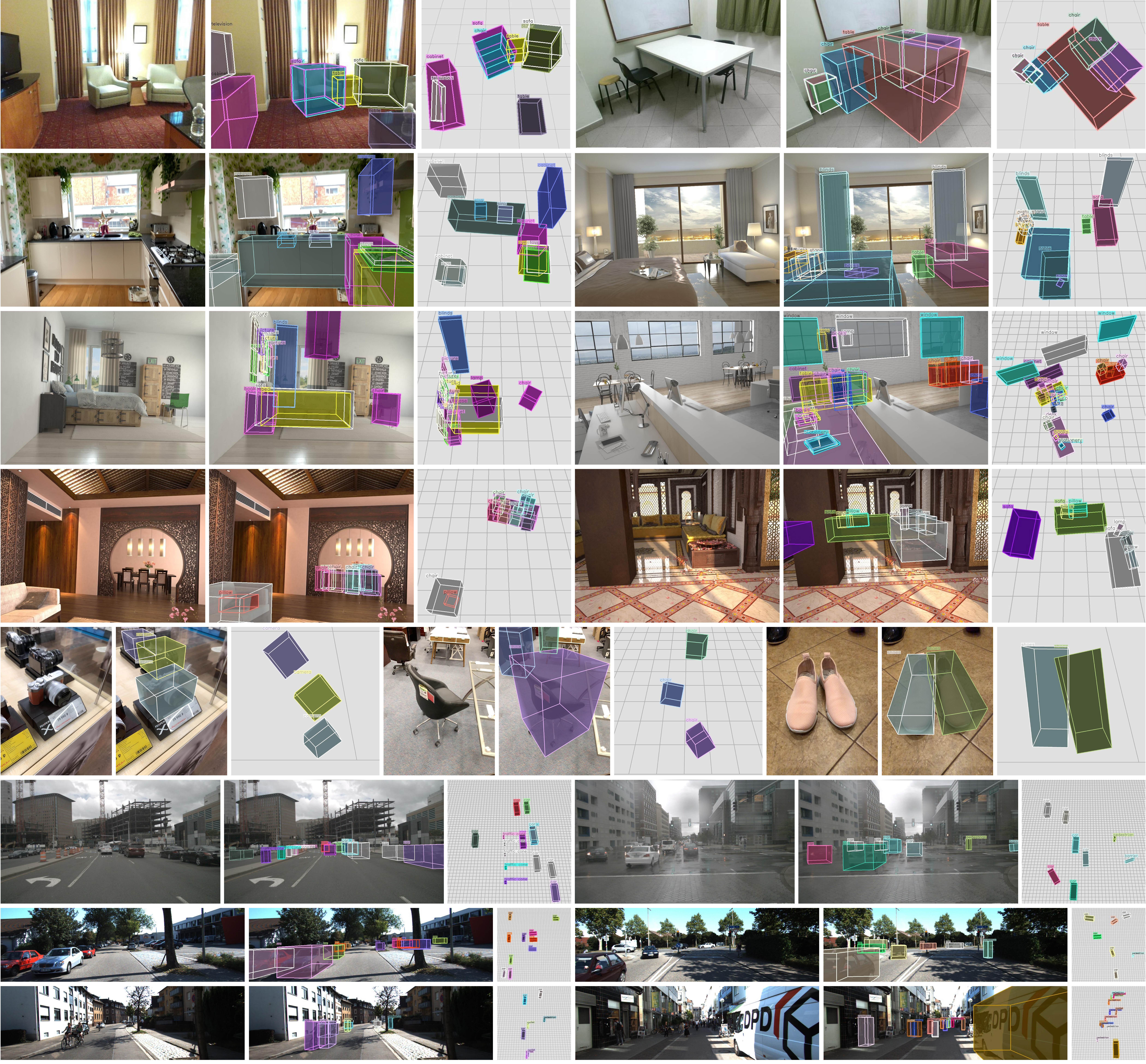}
\vspace{-2mm}
\caption{\method on \dataset test. We show the input image, the 3D predictions overlaid on the image and a top view. 
We show examples from SUN RGB-D~\cite{song2015sun}, ARKitScenes~\cite{dehghan2021arkitscenes}, Hypersim~\cite{hypersim}, Objectron~\cite{objectron2021}, nuScenes~\cite{caesar2020nuscenes}, and KITTI~\cite{Geiger2012CVPR}.}
\label{fig:qualitative_suppl}
\vspace{-4mm}
\end{figure*}

\begin{figure*}[t]
\centering
\includegraphics[width=0.99\linewidth]{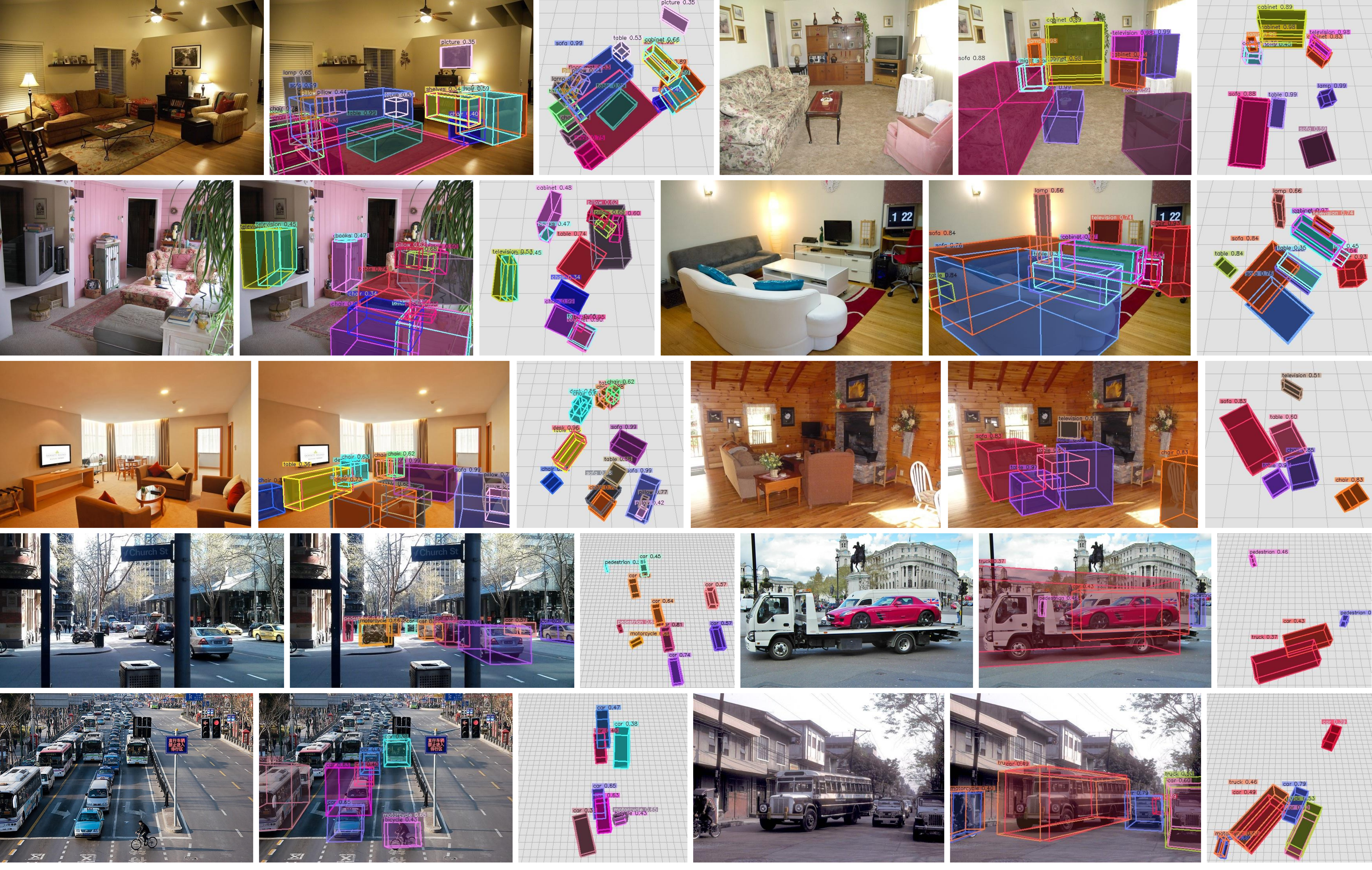}
\caption{\method on COCO in the wild images. We select interesting scenes and observe that generalization performs well for indoor (Rows 1-3) compared to outdoor (Rows 4-5), which appear to fail when in the wild objects or cameras have high variation in unseen poses. }
\label{fig:qualitative_coco}
\end{figure*}

Figure~\ref{fig:qualitative_suppl} shows more \method predictions on \dataset test.
In Figure~\ref{fig:qualitative_coco}, we demonstrate generalization for interesting scenes in the wild from COCO~\cite{lin2014microsoft} images. 
When projecting on images with unknown camera intrinsics, as is the case for COCO, we visualize with intrinsics of $f=2\cdot H,~p_x=\frac{1}{2}W,~p_y=\frac{1}{2}H$, where $H\times W$ is the input image resolution.
As shown in Figure~\ref{fig:qualitative_coco}, this appears to result in fairly stable generalization for indoor and more common failure cases concerning unseen object or camera poses in outdoor. 
We note that simple assumptions of intrinsics prevent our 3D localization predictions from being real-world up to a scaling factor. 
This could be resolved using either real intrinsics or partially handled via image-based self-calibration~\cite{hemayed2003survey, itu2020self, zhuang2019degeneracy} which is itself a challenging problem in computer vision.

Lastly, we provide a demo video\footnote{\url{https://omni3d.garrickbrazil.com/\#demo}} which further demonstrates \method's generalization when trained on \dataset and then applied to in the wild video scenes from a headset mounted camera, similar to AR/VR heasets. 
In the video, we apply a simple tracking algorithm which merges predictions in 3D space using 3D IoU and category cosine similarity when comparing boxes. 
We show the image-based boxes on the left and a static view of the room on the right. 
We emphasize that this video demonstrates \textbf{zero-shot performance} since no fine-tuning was done on the domain data.

{\small
\bibliographystyle{ieee_fullname}
\bibliography{references}
}

\end{document}